\documentclass[sigconf]{acmart}
\usepackage{libertine}

\AtBeginDocument{%
  \providecommand\BibTeX{{%
    \normalfont B\kern-0.5em{\scshape i\kern-0.25em b}\kern-0.8em\TeX}}}

\setcopyright{acmcopyright}
\copyrightyear{2018}
\acmYear{2018}
\acmDOI{10.1145/1122445.1122456}

\acmConference[Woodstock '18]{Woodstock '18: ACM Symposium on Neural
  Gaze Detection}{June 03--05, 2018}{Woodstock, NY}
\acmBooktitle{Woodstock '18: ACM Symposium on Neural Gaze Detection,
  June 03--05, 2018, Woodstock, NY}
\acmPrice{15.00}
\acmISBN{978-1-4503-XXXX-X/18/06}

\newcommand{\etal}{\textit{et al}.}

\newcommand{\eg}{\textit{e}.\textit{g}.}

\begin{document}

\title{Progressive Depth Learning for Single Image Dehazing}



\author{Yudong Liang}
\email{liangyudong@sxu.edu.cn}
\affiliation{%
	\institution{Shanxi University}
	\streetaddress{Wucheng Road}
	\city{Taiyuan}
	\state{Shanxi}
    \country{China}
	\postcode{030006}
}

\author{Bin Wang}
\affiliation{%
	\institution{Shanxi University}
	\streetaddress{Wucheng Road}
	\city{Taiyuan}
	\state{Shanxi}
     \country{China}
	\postcode{030006}
}
\author{Jiaying Liu}
\affiliation{%
  \institution{The Institute of Computer Science and Technology, Peking University, Beijing, China}
}

\author{Deyu Li}
\affiliation{%
	\institution{Shanxi University}
	\streetaddress{Wucheng Road}
	\city{Taiyuan}
	\state{Shanxi}
    \country{China}
	\postcode{030006}
}

\author{Sanping Zhou}
\affiliation{%
 \institution{Institute of Artificial Intelligence and Robotics, Xi'an Jiaotong University, China.}
 }

\author{Wenqi Ren}
\affiliation{%
  \institution{State Key Laboratory of Information Security, Institute of Information Engineering, Chinese Academy of Sciences.}
}


\begin{abstract}
The formulation of the hazy image is mainly dominated by the reflected lights and ambient airlight. Existing dehazing methods often ignore the depth cues and fail in distant areas where heavier haze disturbs the visibility. However, we note that the guidance of the depth information for transmission estimation could remedy the decreased visibility as distances increase. In turn, the good transmission estimation could facilitate the depth estimation for hazy images. In this paper, a deep end-to-end model that iteratively estimates image depths and transmission maps is proposed to perform an effective depth prediction for hazy images and improve the dehazing performance with the guidance of depth information. The image depth and transmission map are progressively refined to better restore the dehazed image. Our approach benefits from explicitly modeling the inner relationship of image depth and transmission map, which is especially effective for distant hazy areas. Extensive results on the benchmarks demonstrate that our proposed network performs favorably against the state-of-the-art dehazing methods in terms of depth estimation and haze removal.
\end{abstract}




\begin{teaserfigure}
\setlength{\tabcolsep}{1pt}
\centering
\footnotesize
\resizebox{1.0\linewidth}{!}
{
\centering
\begin{tabular}{ccc}
\includegraphics[width=0.33\linewidth]{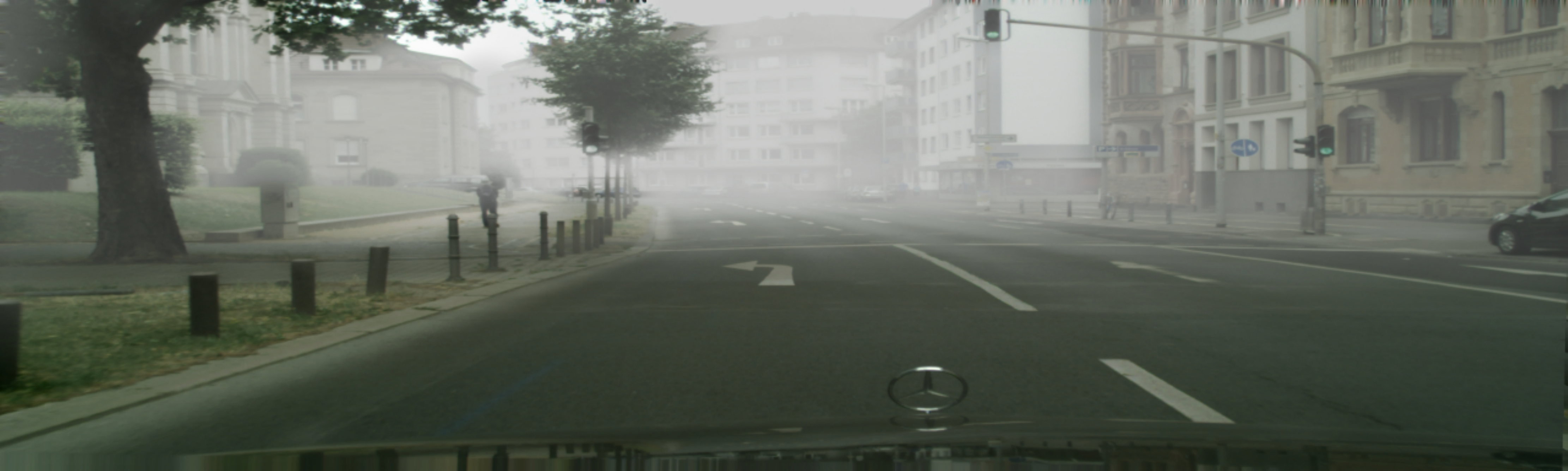}&
\includegraphics[width=0.33\linewidth]{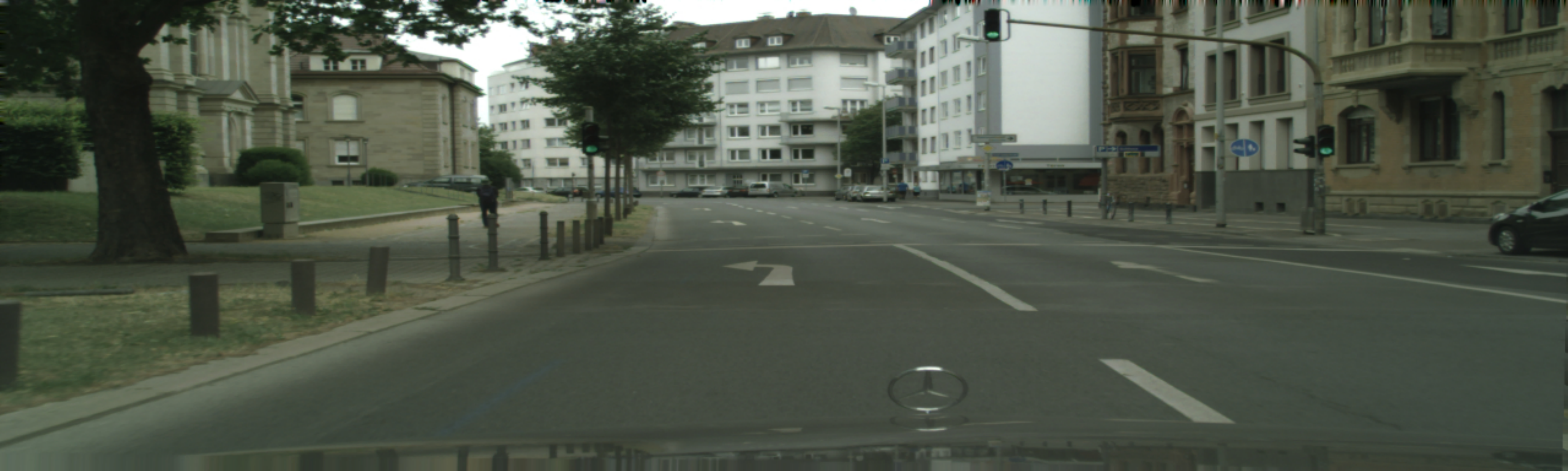}&
\includegraphics[width=0.33\linewidth]{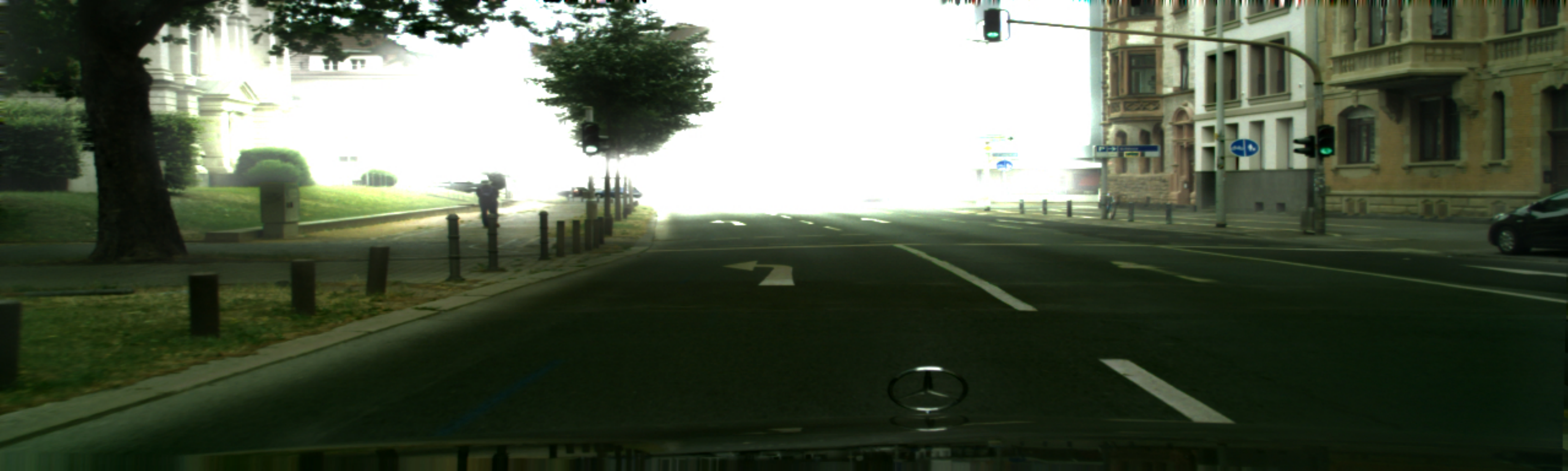}\\
Hazy input & Ground truth & DCP~\cite{he2010single} \\
\includegraphics[width=0.33\linewidth]{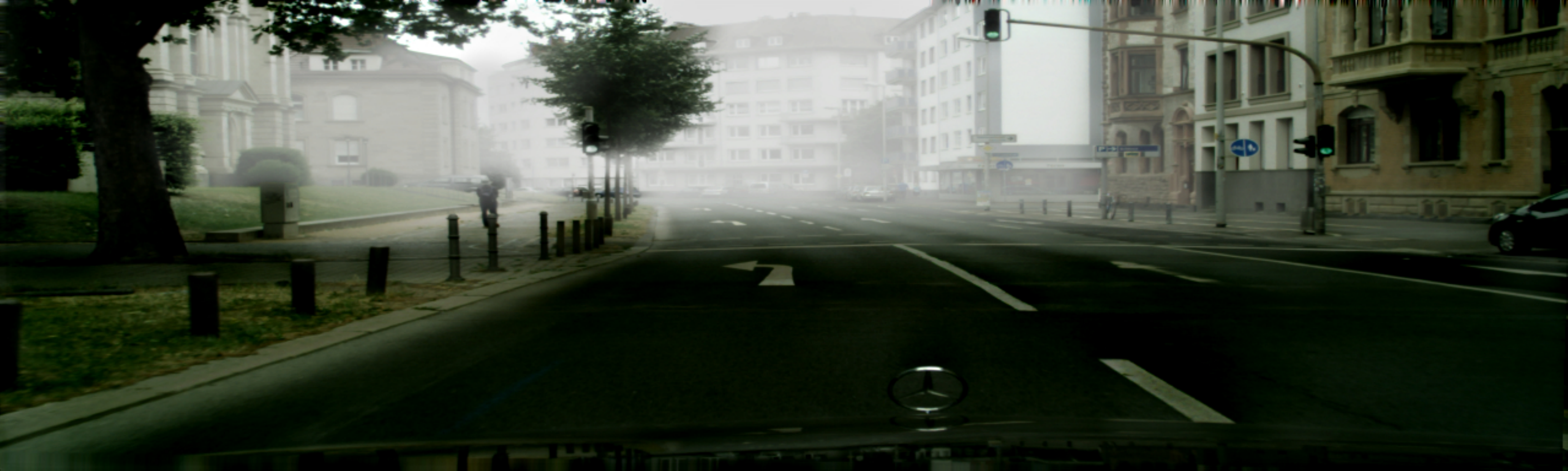}&
\includegraphics[width=0.33\linewidth]{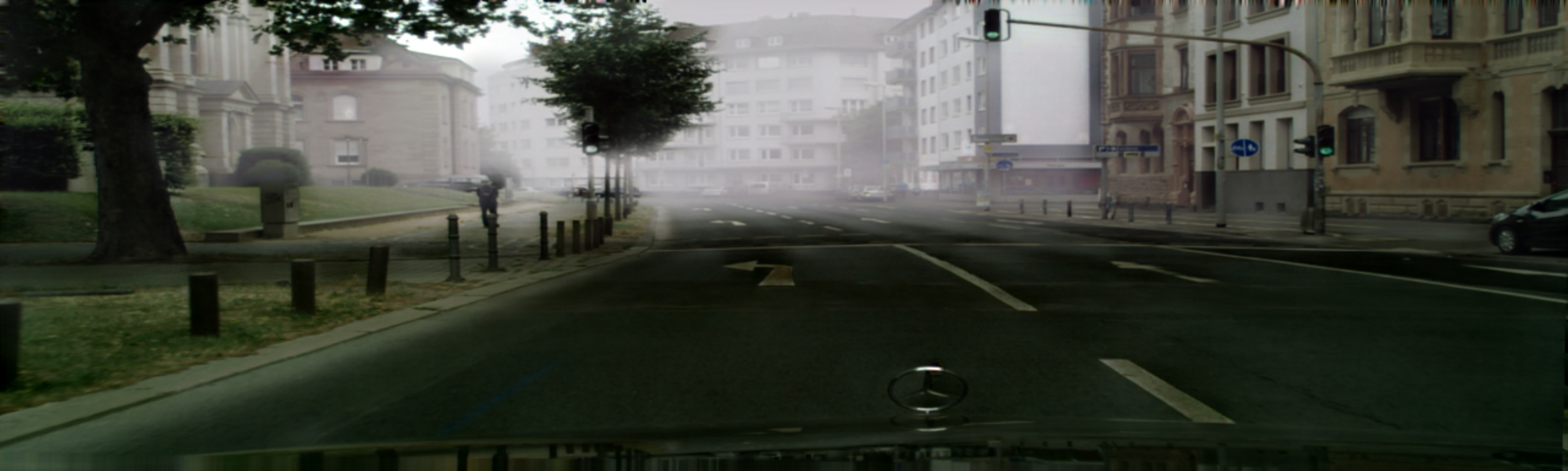}&
\includegraphics[width=0.33\linewidth]{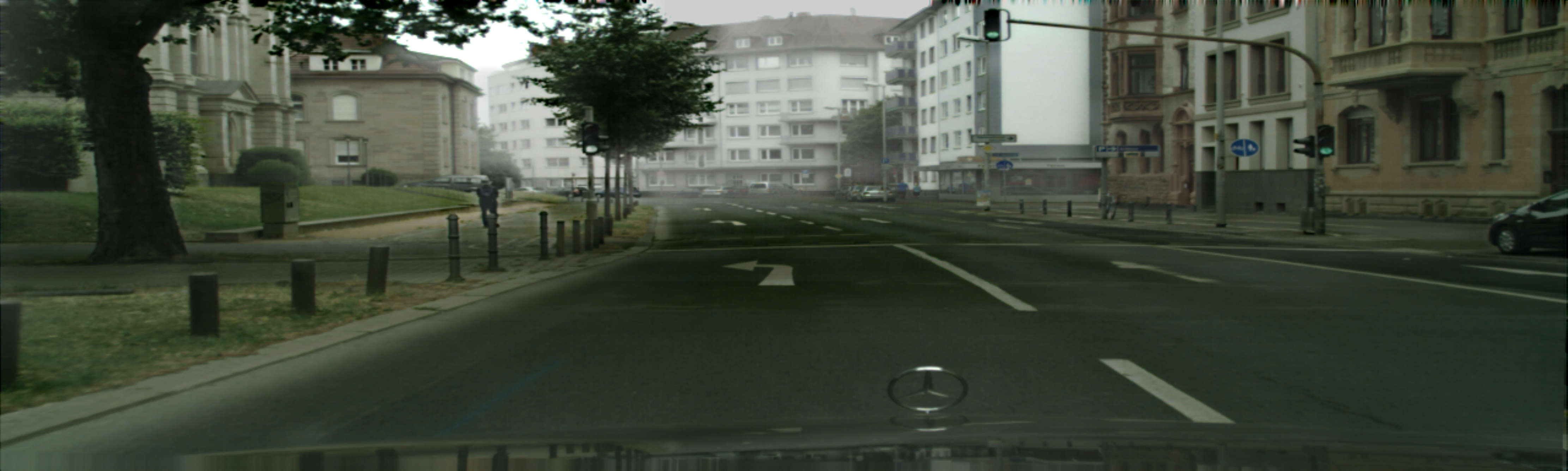}\\
 DCPDN~\cite{zhang2018densely}  & GCANet~\cite{chen2018gated}  & PDLD(ours) \\
\end{tabular}
}
\caption{Visual comparisons for different dehazing methods.}
\label{fig:overview}
\end{teaserfigure}

\maketitle

\section{Introduction}
\label{sec:intro}

Image dehazing is a hot topic with high practical value since the haze disturbs the human observations and impairs the performance of vision-based algorithms. The atmospheric scattering model \cite{israel1959koschmieders} is the widely applied model to describe the formulation of a hazy image as follow
 \begin{equation}\label{eq:scatter}
    \begin{aligned}
    I(x)=J(x)t(x)+A(1-t(x)),
    \end{aligned}
 \end{equation}
where $x$ denotes the location of the pixel, $I(x)$ is the observed hazy image, $J(x)$ is the hazy-free image, A is the global atmospheric light which is supposed to be homogeneous.
In addition, $t(x)=e^{-\beta d(x)}$ denotes the transmission map which gives the portion of the light that reaches the camera, $\beta$ denotes the scattering coefficient of the atmosphere. When the atmosphere is homogeneous, the transmission decays exponentially with the scene depth $d(x)$ in the medium. Existing dehazing methods often fail in distant areas where heavier hazes disturb the visibility.

\begin{figure}[t]
  \centering
  \includegraphics[width=\linewidth]{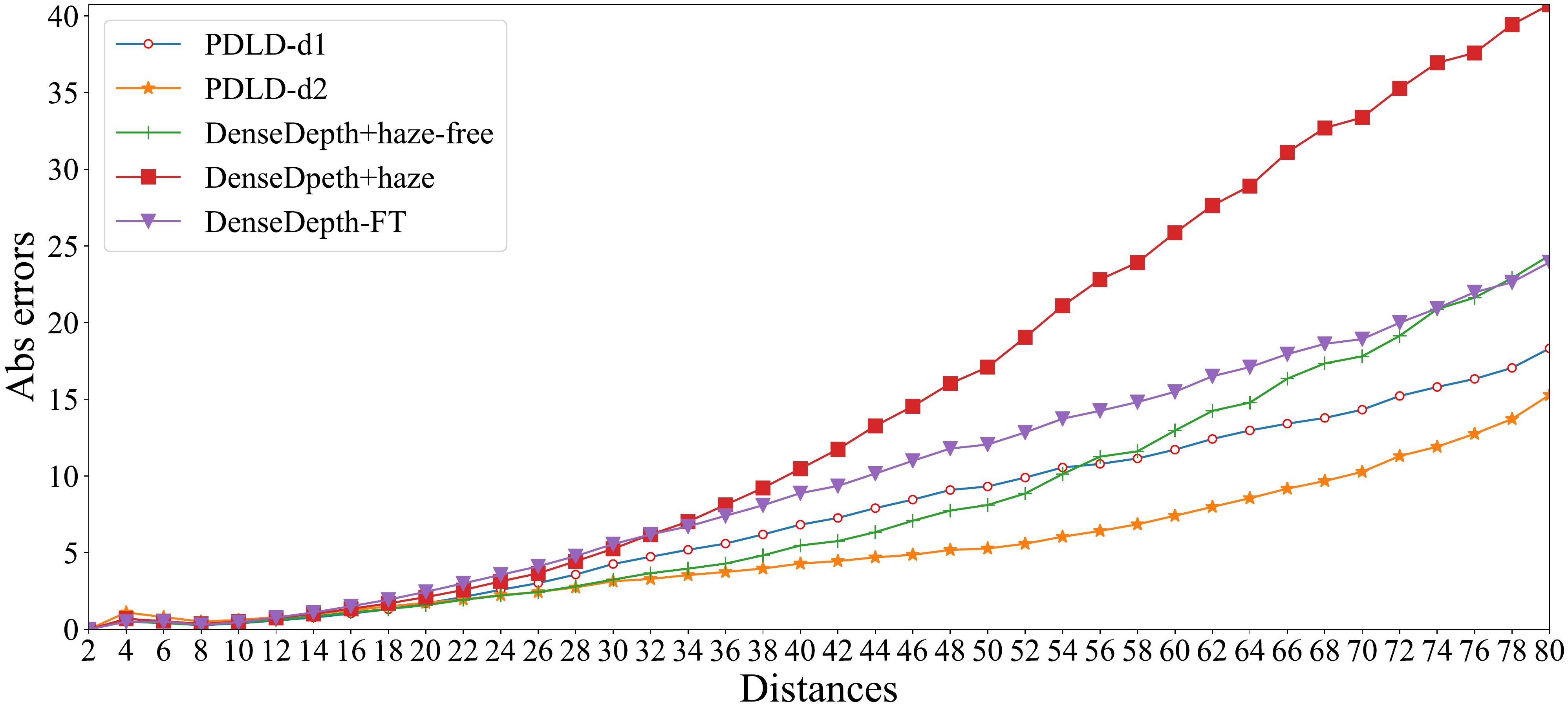}
  \caption{Comparisons of average abs errors for depth estimations to hazy and hazy-free images by deep models in KITTI dataset. DenseDepth\cite{Alhashim2018} is a state-of-the-art model, DenseDepth-FT denotes the model finetuned for hazy images, PDLD-d1 and PDLD-d2 denote the depth estimation modules in the first and second stage respectively.}
  \label{fig:comDepth}
\end{figure}

Accurate estimations for transmission maps are critical to good reconstructions for dehazed images~\cite{cai2016dehazenet,ren2016single,zhang2018densely}. The estimation of the transmission maps from hazy images is highly under-constrained. It is intuitive to exploit image depth information to guide the reconstruction process of the transmission map. However, the image depth is not always available, and predicting depth for a hazy image is also hard. Although there are large amounts of work for monocular (single) or binocular image depth estimation problems, most of them focus on the images captured in clear weather. In outdoor applications, depth estimation algorithms may fail in extreme weather such as haze or fog. The features in hazy images, especially the distant regions, are less distinctive due to the decreased visibility. In Fig.\ref{fig:comDepth}, average pixel-wise abs errors of several approaches are investigated for the pixels whose depths are within certain distances range $[d-2,d]$ in the horizontal axis in KITTI dataset. `DenseDepth+hazy-free / hazy' denotes a state-of-the-art depth estimation model DenseDepth~\cite{Alhashim2018}  performs on the hazy-free / hazy images. It is obvious that the performance of the existed state-of-the-art depth estimation model is severely disturbed by the haze, especially for the distant areas. To firstly execute dehazing and then do depth estimation, or to fine-tune the CNNs with hazy images, is reported to be marginally helpful for stereo matching task~\cite{song2020simultaneous}. 
Depth estimations for the distant area is more difficult than for the nearby regions. Please refer to the experiments part for more details and analysis. On the other hand, high-quality depth maps can be obtained as a by-product of well-estimated transmission maps via haze removal~\cite{he2010single}. Image dehazing and depth estimation become a chicken-and-egg problem.

Inspired by the significant progress of transfer learning and the recent deep models~\cite{Alhashim2018,bhoi2019monocular} for sharp image depth estimation, depth estimation models for hazy-free images may give a reasonable initialization for guiding the transmission map predictions from hazy images. As a transmission map is obtained, the transmission map could assist in learning the depth information in turn. The insight is that the features such as structures or textures in the hazy-free image are more distinctive than in hazy images. Monocular(single) depth predictions largely rely on capturing structures or textures. Well trained depth estimation models for hazy-free images are able to extract some subtle structures or textures. As hazy-free and hazy image domains are closely related, transfer learning would largely reduce the training difficulties and transfer the domain information and give a good initialization for hazy image depth estimations.

 In this paper, a progressive depth learning approach for single image dehazing is proposed to estimate and refine the depth and transmission map iteratively. An end-to-end deep model is designed with this approach to improve the dehazing process with depth information guidance. The model finally reconstructs the dehazed image with the estimated transmission map and global atmosphere light. Our approach benefits from explicit modeling the inner relationship of image depth information and transmission map. Extensive results demonstrate that our proposed algorithm achieves state-of-the-art dehazing performance in the benchmarks. It is especially effective for the outdoor hazy dataset or distant hazy areas, which are more difficult cases for existing dehazing methods. The performance of depth estimations for haze image is also significantly improved.
 All of these declare the power of progressive depth learning for image dehazing which exploits the inner relationship between image depth estimation and image dehazing.

\section{Related Work}

Single image dehazing and depth estimation are both ill-posed and highly under-constrained problems.

\textbf{Image dehazing} can be roughly classified into handcrafted priors~\cite{tan2008visibility,he2010single} or learning-based methods~\cite{zhang2018densely}. The statistical properties of haze-free images are exploited to refine the hazy images, \eg, dark channel prior~\cite{he2010single} and color-line prior~\cite{fattal2014dehazing}. He \etal~\cite{he2010single} proposed a dark channel prior and figured out a high-quality depth map according to the transmission map.
Instead of developing skillful handcrafted priors, learning-based methods, especially the deep learning-based approaches, learn the mapping from a hazy image to the dehazed output. Cai \etal~\cite{cai2016dehazenet} designed a DehazeNet for the estimation of the transmission map and then followed the atmospheric scattering model to reconstruct the dehazed image. Ren \etal~\cite{ren2016single} combined multi-scale information with a deep network to predict the transmission map. Li \etal~\cite{li2017aod} designed an end-to-end light-weight CNN directly restore clean images from the hazy input. Via adversarial network training, Zhang and Patel~\cite{zhang2018densely} proposed a Densely Connected Pyramid Dehazing Network using an edge-preserving loss termed DCPDN. A joint-discriminator for the corresponding dehazed image and the estimated transmission map are applied to incorporate the mutual structural information between the estimated transmission map and the dehazed result. Although DCPDN has achieved state-of-the-art results in the benchmarks, this method has a high computational burden since the dense connections. Qu \etal~\cite{qu2019enhanced} treated the dehazing problem as a pixel to pixel image translation problem and directly restored the clear image. The deep learning-based methods~\cite{cai2016dehazenet,zhang2018densely,Shao2020Domian} have largely improved the dehazing performances since deep networks construct more sophisticate features than handcrafted priors based methods. However, the depth information is ignored in the inference of transmission map during the dehazing process by the existing deep learning-based approaches, which may hinder the dehazing performance in the distant areas. Some artifacts like halos may appear in the dehazed images.

\begin{figure*}[ht]
     \centering
     \footnotesize
    \includegraphics[width=\textwidth]{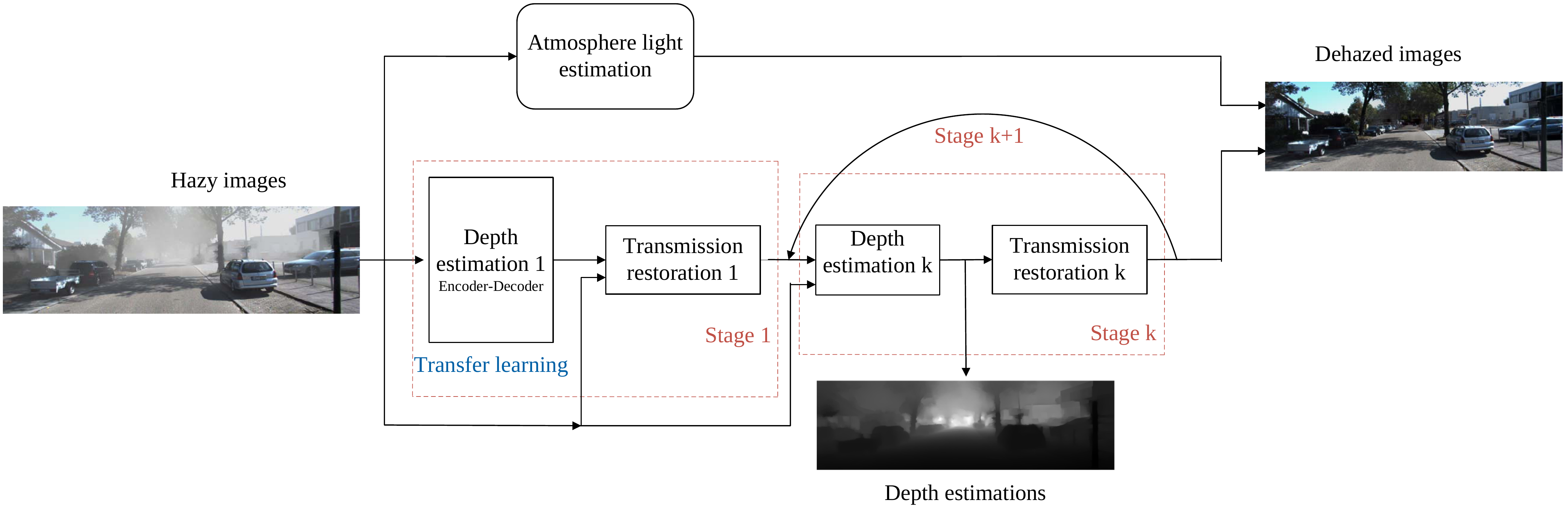}
    \caption{The framework of the proposed progressive learning approach, the depth and transmissions are iteratively estimated in a a circular and stage-wise cascade network, then the transmissions are combined with the predicted atmosphere light to restore the dehazed images following the atmosphere scattering model.}
  \label{fig:framework}
  \end{figure*}

\textbf{Depth estimation} methods recover depth information from monocular (single) or binocular images~\cite{Alhashim2018,bhoi2019monocular}. Alhashim and Wonka~\cite{Alhashim2018} transferred a dense network pre-trained by image classification task to image depth estimation, which achieved the state-of-the-art performance among the monocular depth estimation task. It indicates some related domain information is beneficial to the depth estimation task. Zhang \etal~\cite{zhang2019ga} proposed a semi-global aggregation layer and the local guided aggregation layer for the stereo matching task. Although there are large amounts of work for monocular (single) or binocular image depth estimation, most of them focus on the images captured in clear weather. Depth estimation is especially important for outdoor applications such as autonomous driving. In outdoor applications, algorithms may fail in the extreme weather such as haze or fog.

Until very recent time, Song \etal~\cite{song2020simultaneous} simultaneously performed the stereo matching and image dehazing in a feature fusion and multitask learning manner, which implicitly models the inner relationship between depth and transmission map. The stereo matching estimates the disparities from a pair of left and right cameras, and the depth is then calculated from the estimated disparities. Although the performances of both the depth estimation and the image dehazing are improved, applying an attention feature fusion may not fully benefit from the inner relationship between depth and transmission map. In contrast, we focus on dehazing and depth estimation from the single image. Our approach explicitly models the relationship between the depth and transmission map, which performs progressive depth learning and iteratively refines the estimation.

\section{Proposed method}
Our proposed approach performs the depth prediction and follows the atmospheric scattering model to restore dehazed images. The main contribution is to exploit the relationship between depth estimation and transmission. As shown in Fig.\ref{fig:framework}, the backbone is an end-to-end deep network that first splits into two branches that estimate transmission map guided by depth and atmosphere light separately, then merges to restore dehazed images. In the training phase, the ground truth depth and haze-free images are referenced for progressive supervised learning. While in the test phase, the model receives hazy images and restores dehazed images as well as depth information. The details are explained as follows.

\subsection{Progressive Learning Network}

As explained in Section \ref{sec:intro}, estimating the depth information and the transmission map are closely related tasks, and both are ill-posed problems. To address this problem, depth and transmission estimation network is designed as a circular and stage-wise cascade network composed of depth estimation module and transmission reconstruction module (with depth guidance) repeatedly, as shown in Fig. \ref{fig:framework}. The loss function for the whole framework is as follows:
 \begin{equation}\label{eq:loss}
   \begin{aligned}
  L=\sum\limits_{k}^{K} (L^k_{d}+L^k_t)+L_a+L_{Dhaze},
 \end{aligned}
\end{equation}
where $L^k_{d}$ and $L^k_t$ denote the loss functions for the depth and transmission estimation in the $k$-th stage respectively, $K$ is the number of the total stages for estimation, $L_a$ denotes the loss function for the atmosphere light estimation and $L_{Dhaze}$ denotes the loss function for the dehazed image. Each term will be introduced in the following parts.

\subsection{Depth and Transmission Estimation}

\textbf{Depth estimation module} in the first stage applies a simple but effective encoder-decoder architecture proposed by \cite{Alhashim2018} for preliminary depth estimation. The encoder is the DenseNet-169 network which encodes the input into a feature vector. The decoder consists of several blocks with $2x$ bilinear up-sampling layers and convolutional layers followed by LeakyRelu to restore the final depth map at half the input resolution.

Following \cite{Alhashim2018}, a weighted combination loss is applied as the loss function for depth estimation in the first stage:
%
\begin{equation}\label{eq:lossDepth}
   \begin{aligned}
  L_{d}(\hat{d}_{k}(x),d)=& \lambda L_{depth}(\hat{d}_{k}(x),d(x))+L_{grad}(\hat{d}_{k}(x),d(x))\\ & +L_{SSIM}(\hat{d}_{k}(x),d(x))
 \end{aligned}
\end{equation}
where $\hat{d}_{k}(x)$ is the reciprocal of the predicted depth in the $k$-th stage for pixel $x$ and $d$ denotes the reciprocal of the ground truth depth, $n$ is the total number of the depth image. Using the reciprocal makes the training numeric more stable and alleviates the problem that loss $L_{d}$ will be larger when the ground truth depth values are bigger.
 $L_{depth}$ denotes the pixel-wise difference between the $\hat{d}_{k}(x)$ and the ground truth $d(x)$, L1 loss can be applied as:
\begin{equation}\label{eq:maeDepth}
    L_{depth}(\hat{d}_{k}(x),d(x))=\frac{1}{n}\sum\limits_{x}^{n}  | \hat{d}_{k}(x)-d(x) |.
\end{equation}
The second term $ L_{grad}$ measures distortions of high-frequency details. The reconstruction errors of the image gradients in the horizontal and vertical directions are denoted as $g_h$ and $g_v$ respectively:
\begin{equation}\label{eq:maeDGrad}
    L_{grad}(\hat{d}_{k}(x),d(x))=\frac{1}{n}\sum\limits_{x}^{n} ( | g_h(d_{k}(x),d(x)) |+| g_v(d_{k}(x),d(x)) |  ).
\end{equation}
The third term applies Structural Similarity (SSIM) which is commonly used in the depth estimations to faithfully restore the structures. $L_{SSIM}(\hat{d}_{k}(x),d(x))$ is defined as :
     \begin{equation}\label{eq:SSIM}
    L_{SSIM}(\hat{d}_{k}(x),d(x))=\frac{1-SSIM(\hat{d}_{k}(x),d(x))}{2}.
    \end{equation}
If the depth information is obtained, Eq.~\ref{eq:scatter} indicates only exponential calculation is needed to reconstruct the transmission. However, Eq.~\ref{eq:scatter} is only an approximation to the complex dehazing problem. Besides, the depth estimation would contain errors. Thus, we use a CNN model to mimic the restoration process from the hazy images with the guidance of depth estimations.

\noindent \textbf{Transmission reconstruction module} upsamples the estimated reciprocal of the depth map to the original size, inverses the reciprocal to calculate the depth back, normalizes and then maps the depth to the transmission. Convolutional layers, InstanceNorm layers, and leakyReLU are exploited to regularize the depth. The normalized depth mpas are concatenated with hazy images along the channels into a four-channel image. A $1 \times 1$ pointwise convolution operations are applied and encoder-decoder networks are proposed to predict transmission maps. $L^2$ or $L^1$ loss $L^k_t$ could be applied to optimize the learning process for the transmissions in the $k$-th stage.


\noindent \textbf{Progressive learning.}
The depth estimation module for hazy images is difficult to train from scratch due to decreased visibility. With the help of transfer learning, the depth estimation module in the first stage is initialized by the well-trained model released by \cite{Alhashim2018} which is able to extract some subtle structures or textures information to predict depth. Empirically, transfer learning has largely boosted the depth estimations performance. Training from scratch leads to terrible performances in our experiments.

A stage-wise training strategy and the loss function in Eq.~\ref{eq:loss} is utilized. It means when in the first stage, only the architecture in the first stage is constructed and trained for the dehazing process.
To avoid interference of optimization for depth estimations, transmission and atmospheric light estimation modules are trained using Eq.~\ref{eq:loss}, while the depth estimation module is fixed during training.

After the model learning in the first stage is finished, the depth estimation and the transmission reconstruction modules can be further built for more stages to refine the estimation progressively. The hazy image is further fused with transmission to be fed into the next depth estimation module. Depth estimation from hazy images can be further and largely improved by transmission guidance. There are multiple benefits to build this circular and stage-wise cascade network composed of depth estimation module and transmission reconstruction module (with depth guidance) repeatedly. Firstly, depth or transmission guidance can significantly reduce the difficulties of estimating the transmission or depth. With the aid of transmission, depth estimations in the following stages can be a much easier task, and the estimation model could be much simpler than the depth estimation model in the previous stage. In turn, the same situation applies to the transmission reconstructions. Secondly, a stage-wise cascade network introduces the depth or transmission in each stage to supervise learning. The supervision in the later stages would assist the training in the previous stages. Third, with a progressively refined guidance, the estimations can be further refined.


In our implementations, considering the trade-off between the performances and the computation burdens, only two stages are constructed, and the second stage only contains depth estimation. The second depth estimation module in our implementations applies the same structures as the encoder-decoder structure in the transmissions estimation module in the first stage. In the test phase, the depth estimation by the second stage would guide the transmission estimation and restore the dehazed images.
The transmission estimation module is shared for the first and second stages in the testing phase, which further reduces the parameter numbers.

The experiments prove that the progressive depth learning strategy improves the depth estimations for the hazy images and largely improves the dehazing performances, benefiting from the more accurate depth and transmission estimations. In fact, since the haze provides some cues for the depth estimations, our depth estimation model with hazy images outperforms the state-of-the-art depth estimation model with hazy-free images.

\subsection{Image Dehazing}

\noindent \textbf{Atmospheric light estimation.} In addition to scene transmission map and depth map, we also need to estimate the atmospheric light $A$ for recovering the clear image.
We use an 8-block U-net structure as \cite{zhang2018densely} and suppose the atmospheric light is homogeneous, where the encoder consists of four Conv-BN-Relu blocks, and the decoder comprises symmetric Dconv-BN-Relu block. $L^2$ or $L^1$ loss could be applied to optimize the learning process for the atmospheric light.

With the estimated transmission and atmospheric light, hazy image $I'(x)$ can be computed from the ground-truth haze-free images I(x) as Eq.~\ref{eq:scatter}. The L1 loss between estimated hazy image $I'(x)$ and hazy input image $I(x)$ can be utilized to self-supervise the reconstruction loss.
 \begin{equation}\label{eq:maeDhz}
    {L_{Dhaze}}=| J(x)t(x)+A(1-t(x))-I(x) |
 \end{equation}

\section{Experiments}

In this section, the effectiveness of the proposed approach is demonstrated by various experiments on image dehazing and image depth estimations. Several synthetic datasets and real-world images are utilized to compare our method against several state-of-the-art methods. The compared dehazing methods includes \textbf{DCP}~\cite{he2010single}, \textbf{AOD-Net}~\cite{li2017aod}, \textbf{DCPDN}~\cite{zhang2018densely}, \textbf{GCANet}, a very recent stereo matching and dehazing work~\cite{song2020simultaneous}. All the reported data is tested by the authors' released code or reported results in their paper. Our progressive depth learning for image dehazing is denoted as \textbf{PDLD}.

Our model is implemented by pyTorch with a RTX 2080Ti GPU, and Adam optimization is utilized during the training. The initial learning rate ($lr$) is set to be 0.002, with the help of the transfer learning, the depth estimation module is initialized by the DenseDepth model~\cite{Alhashim2018}, and then the learning rate is set to be $0.1 \times lr$.

Analogous to most of existing deep learning-based dehazing methods~\cite{cai2016dehazenet,ren2016single,li2017aod,zhang2018densely}, dehazing datasets are synthesized for the training and testing process.

\subsection{Indoor Dataset}
The NYU-depth2 dataset \cite{silberman2012indoor} is a widely applied indoor dataset for dehazing comparisons.
Following the common setting \cite{Alhashim2018} of depth estimation task, the NYU-depth2 dataset splits into 120K training images and 654 testing images.

\begin{figure*}
\setlength{\tabcolsep}{1pt}
\centering
\footnotesize
\resizebox{1.0\linewidth}{!}
{
\centering
\begin{tabular}{ccccccc}
\includegraphics[width=0.14\linewidth]{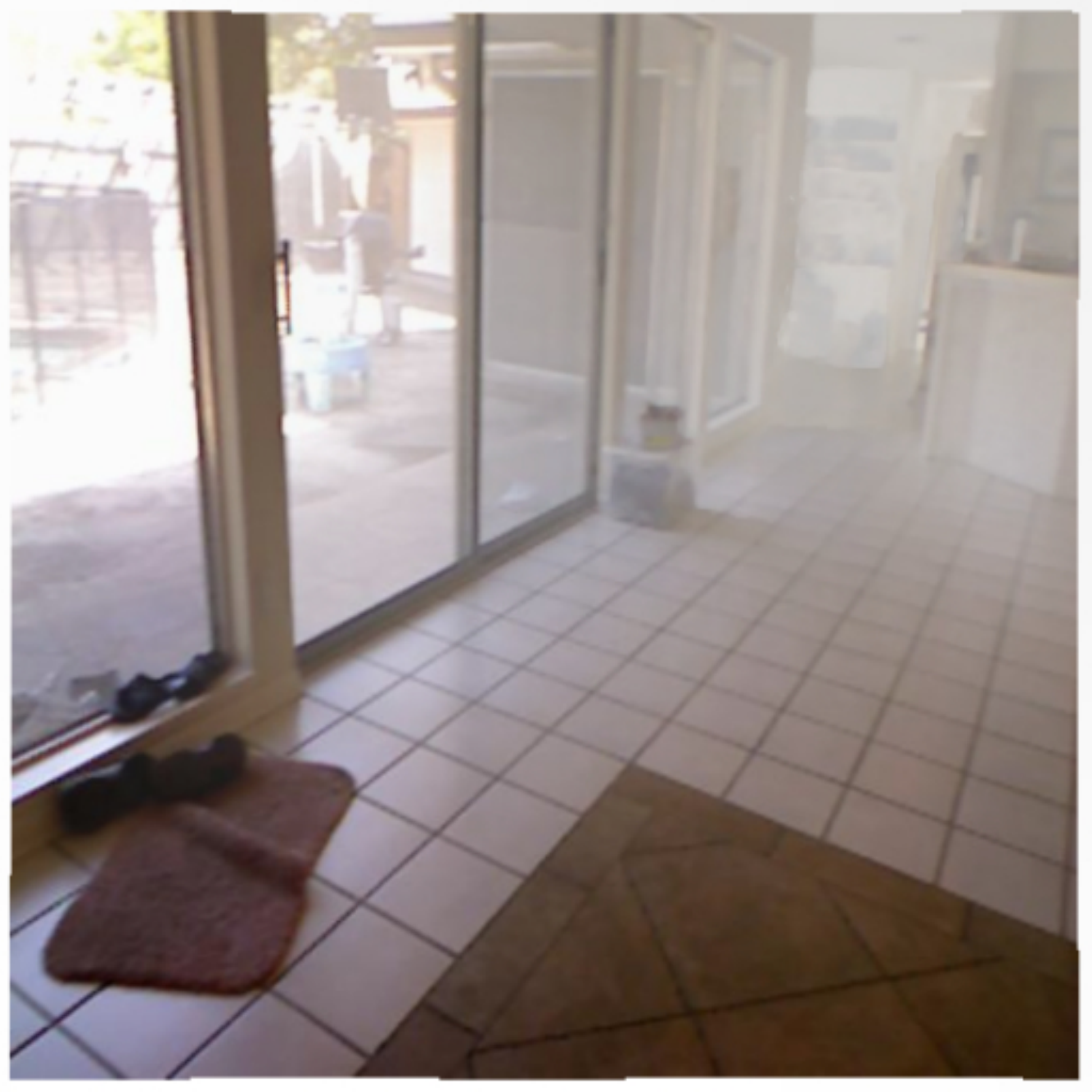}&
\includegraphics[width=0.14\linewidth]{NYUv2/5gt}&
\includegraphics[width=0.14\linewidth]{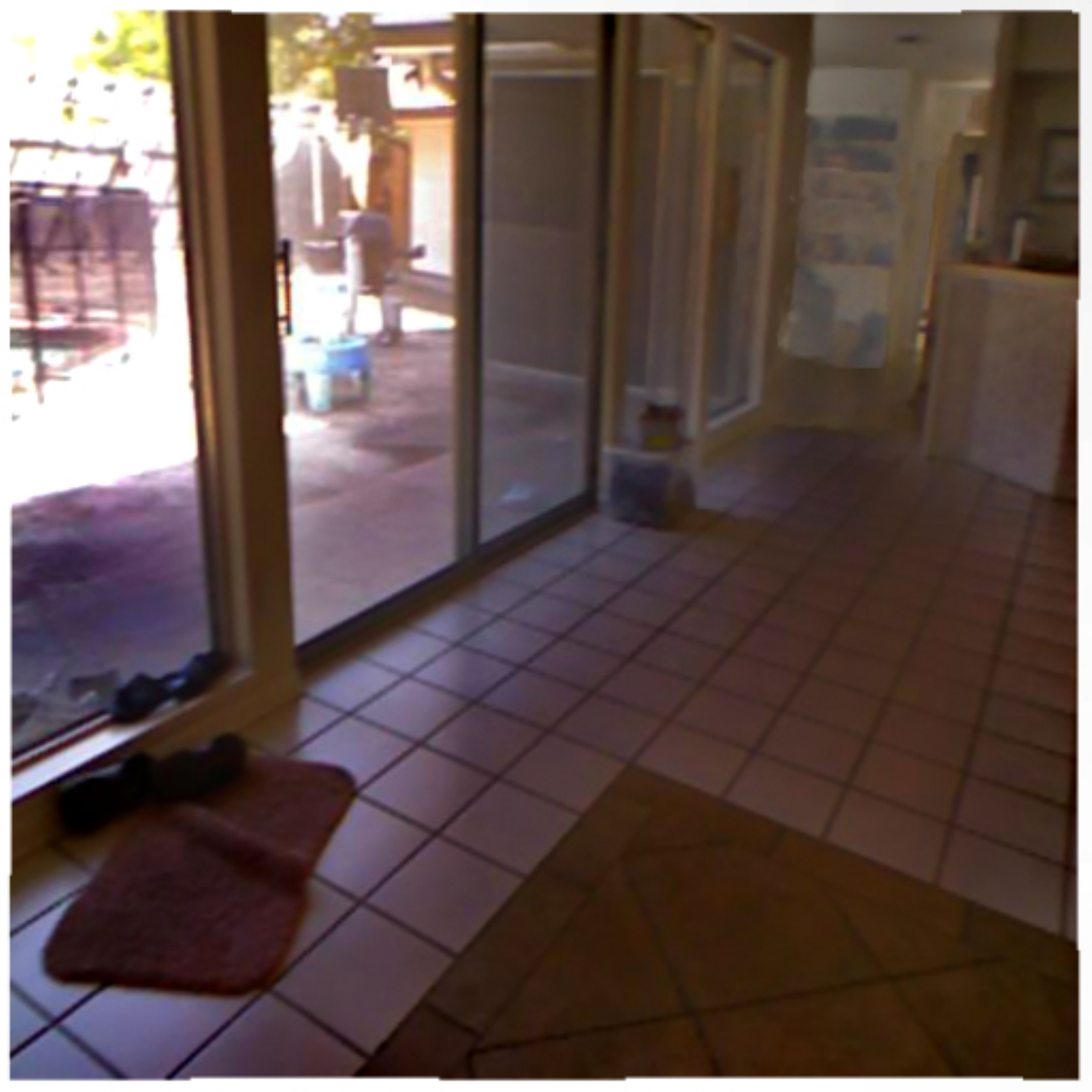}&
\includegraphics[width=0.14\linewidth]{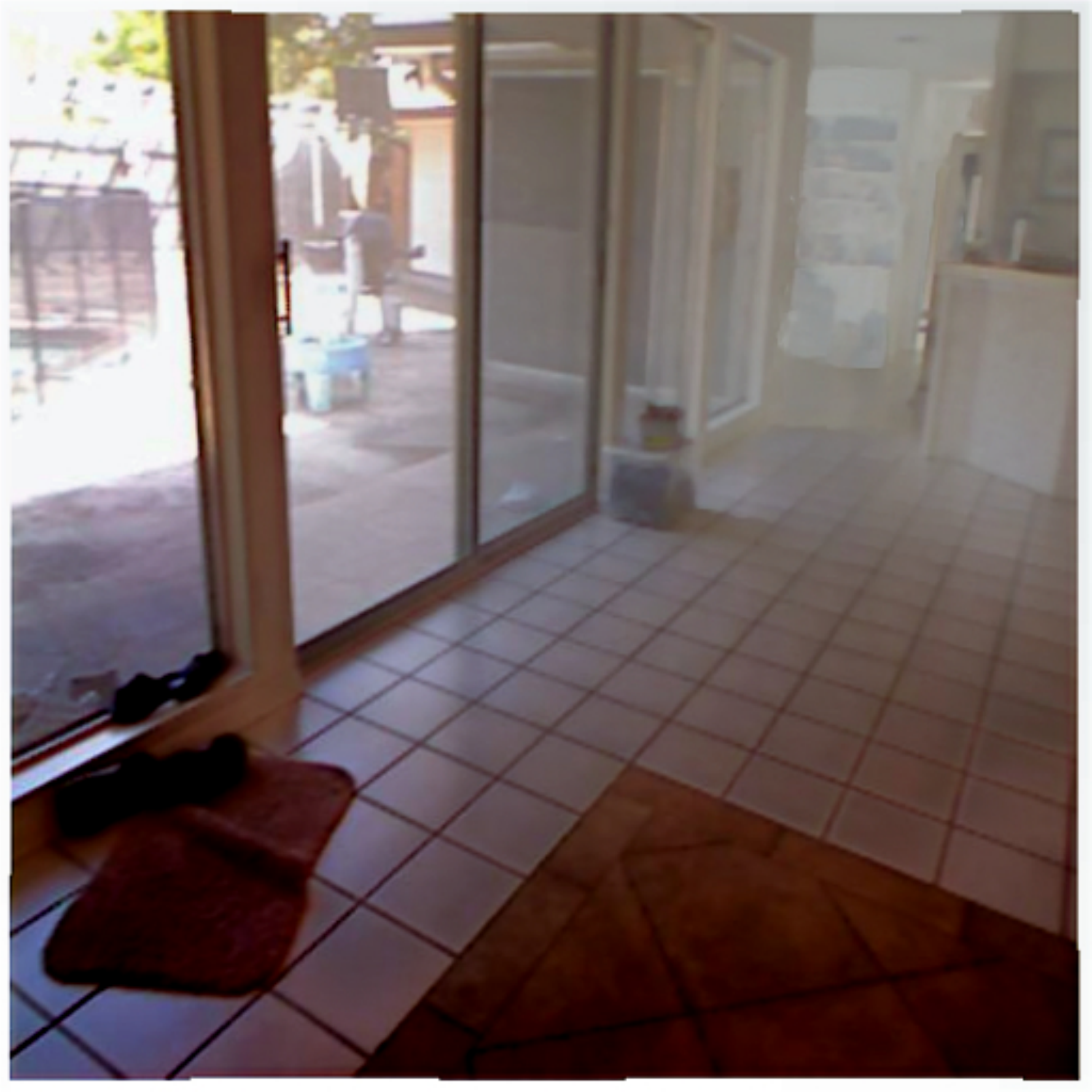}&
\includegraphics[width=0.14\linewidth]{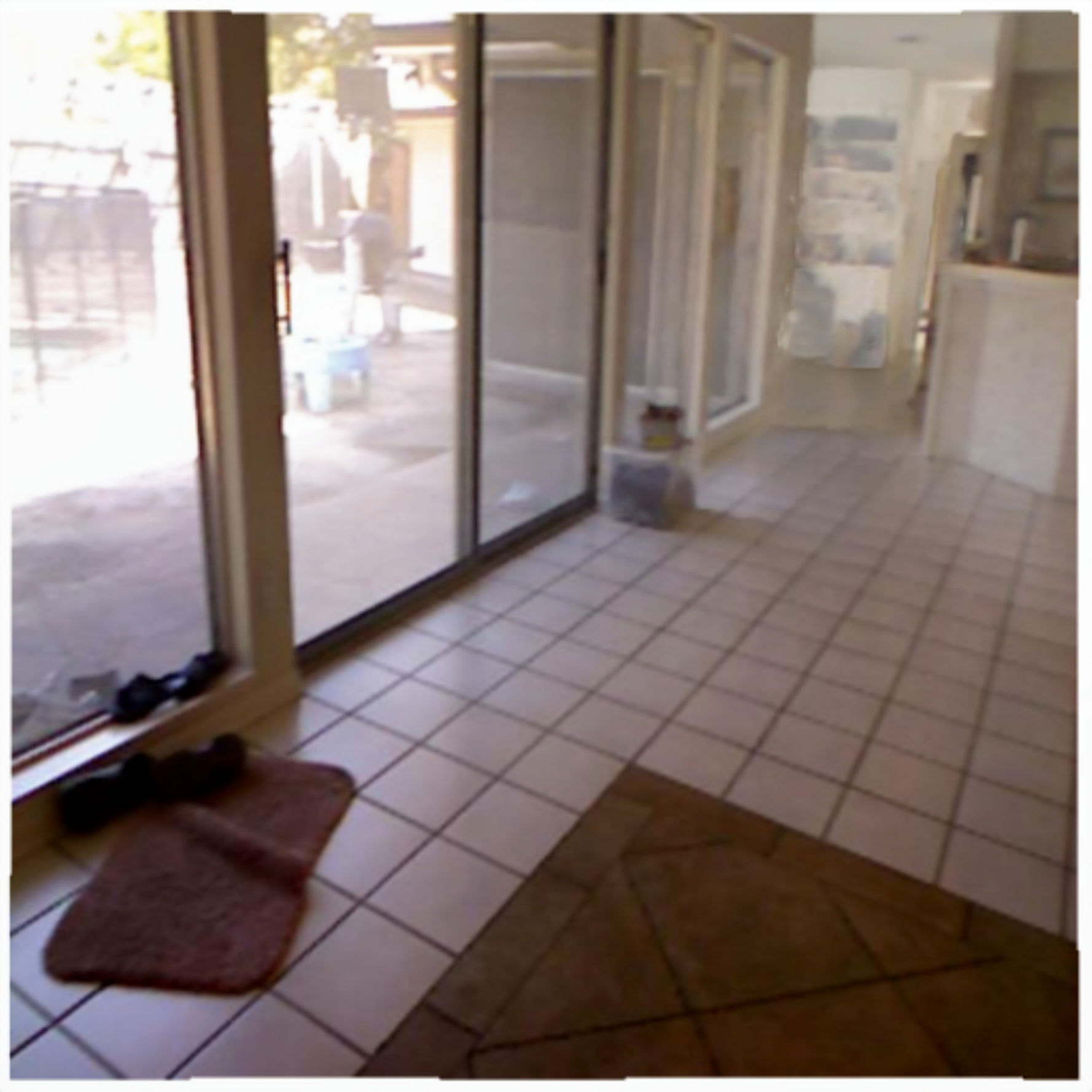}&
\includegraphics[width=0.14\linewidth]{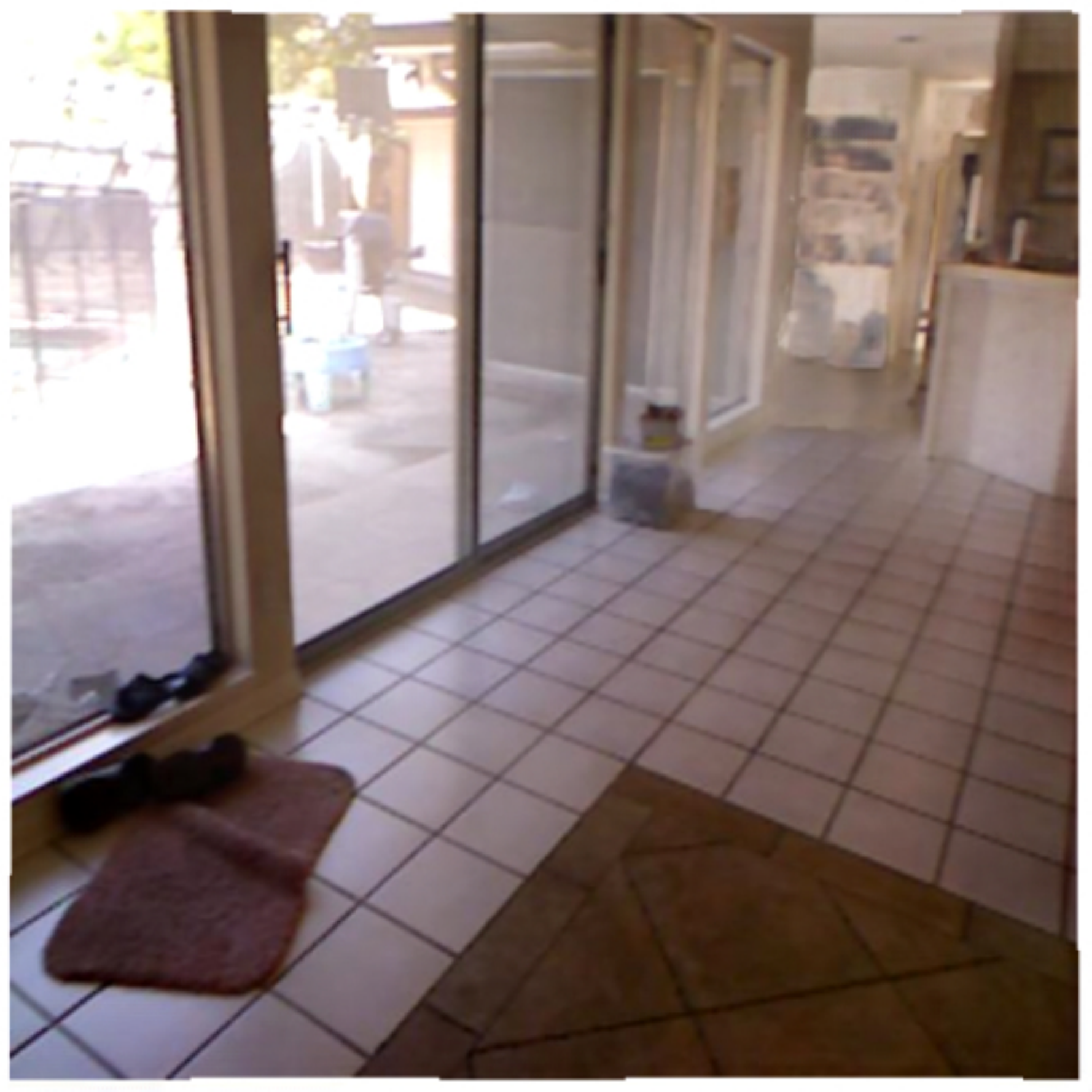}&
\includegraphics[width=0.14\linewidth]{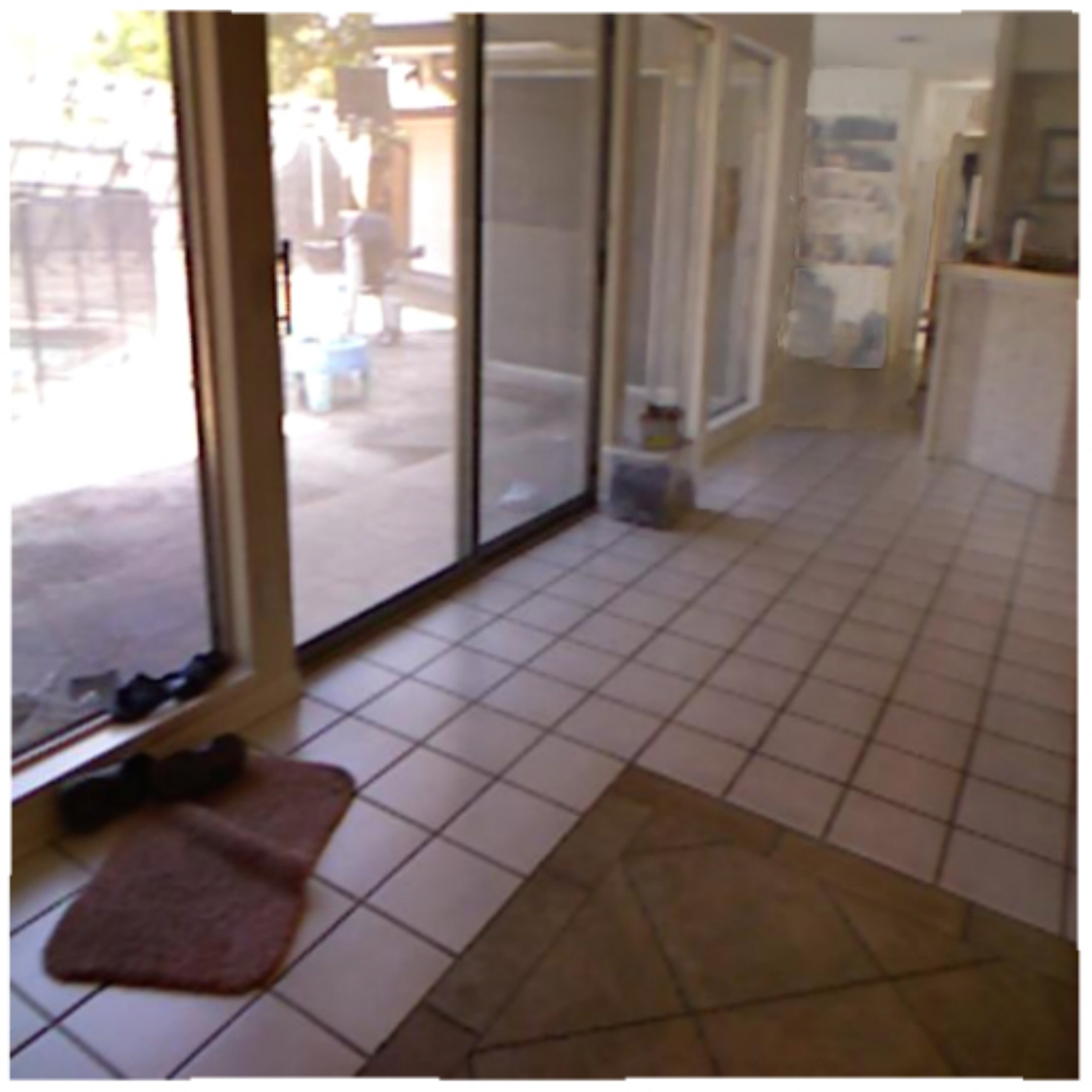} \\
 PSNR / SSIM &  &16.51 / 0.8519&16.83 / 0.8207&19.16 / 0.9328 & 19.46 / 0.9434&21.70 / 0.9593\\
\includegraphics[width=0.14\linewidth]{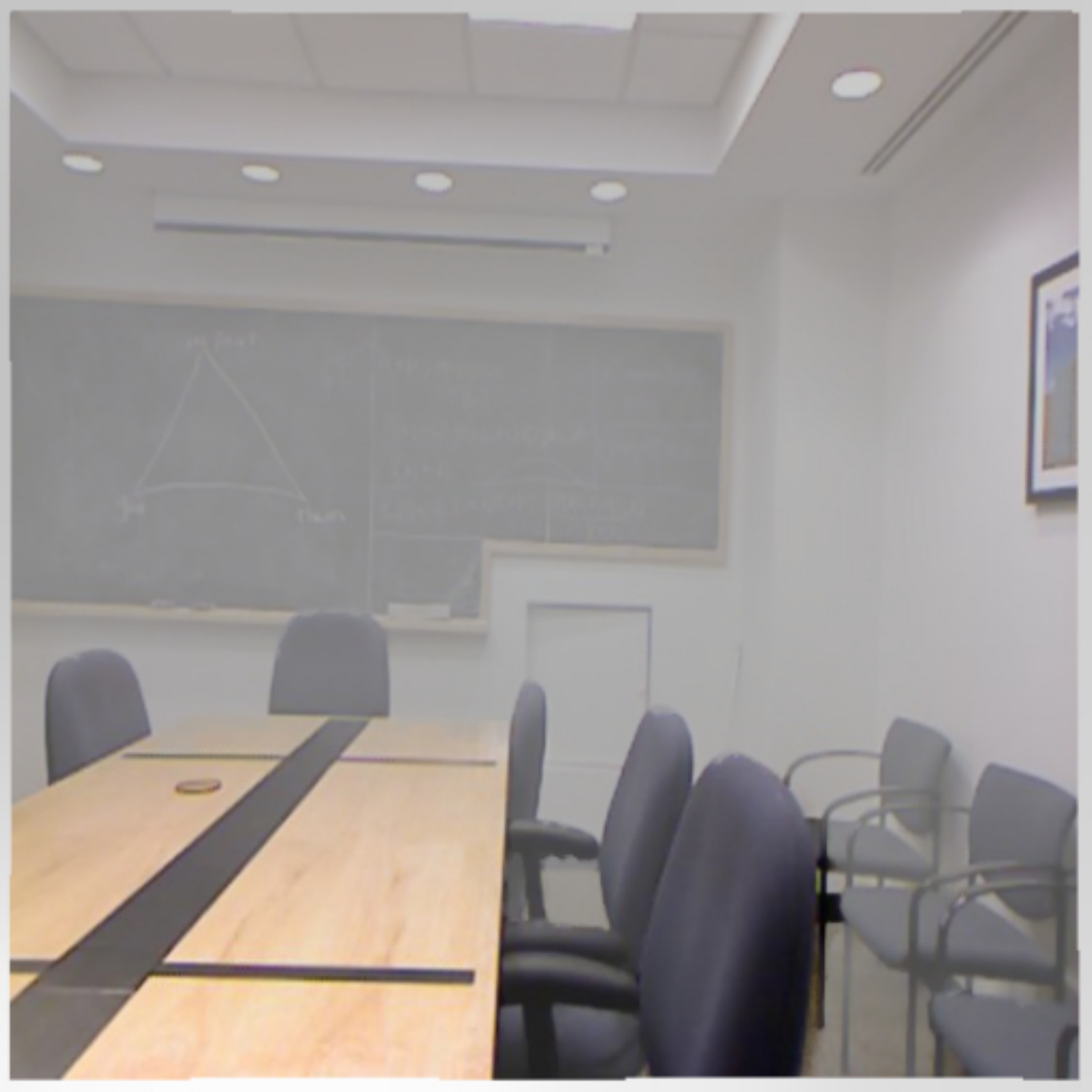}&
\includegraphics[width=0.14\linewidth]{NYUv2/162gt}&
\includegraphics[width=0.14\linewidth]{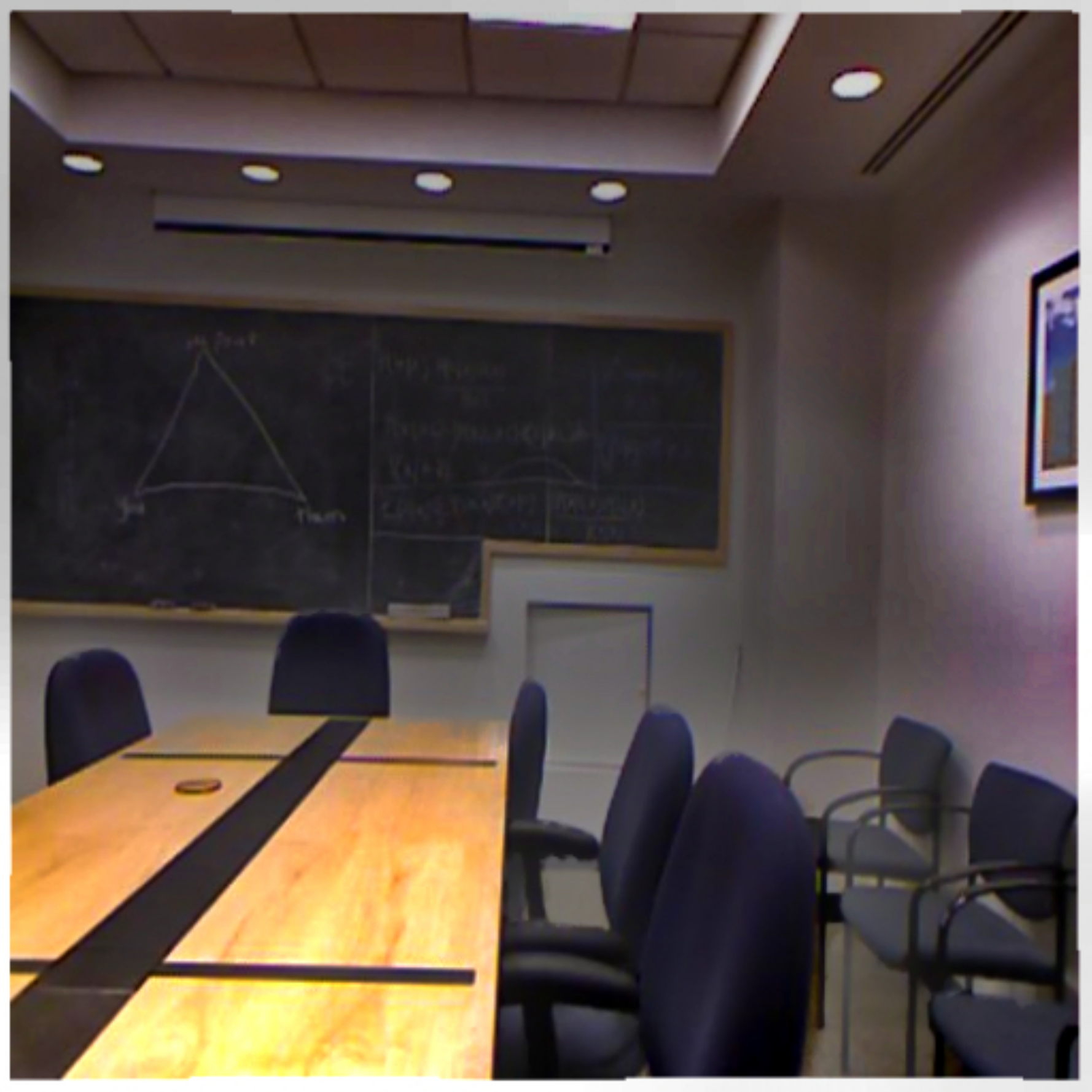}&
\includegraphics[width=0.14\linewidth]{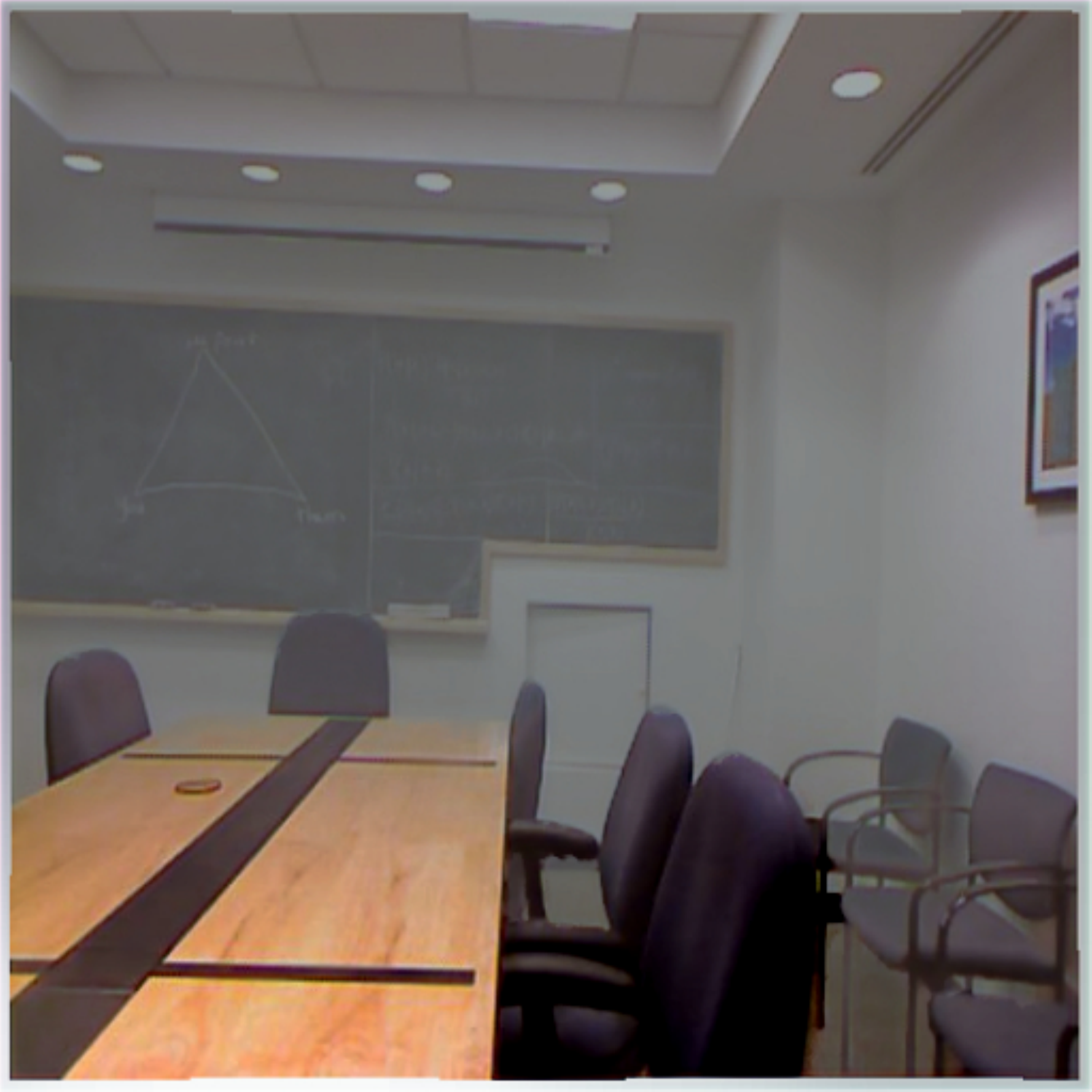}&
\includegraphics[width=0.14\linewidth]{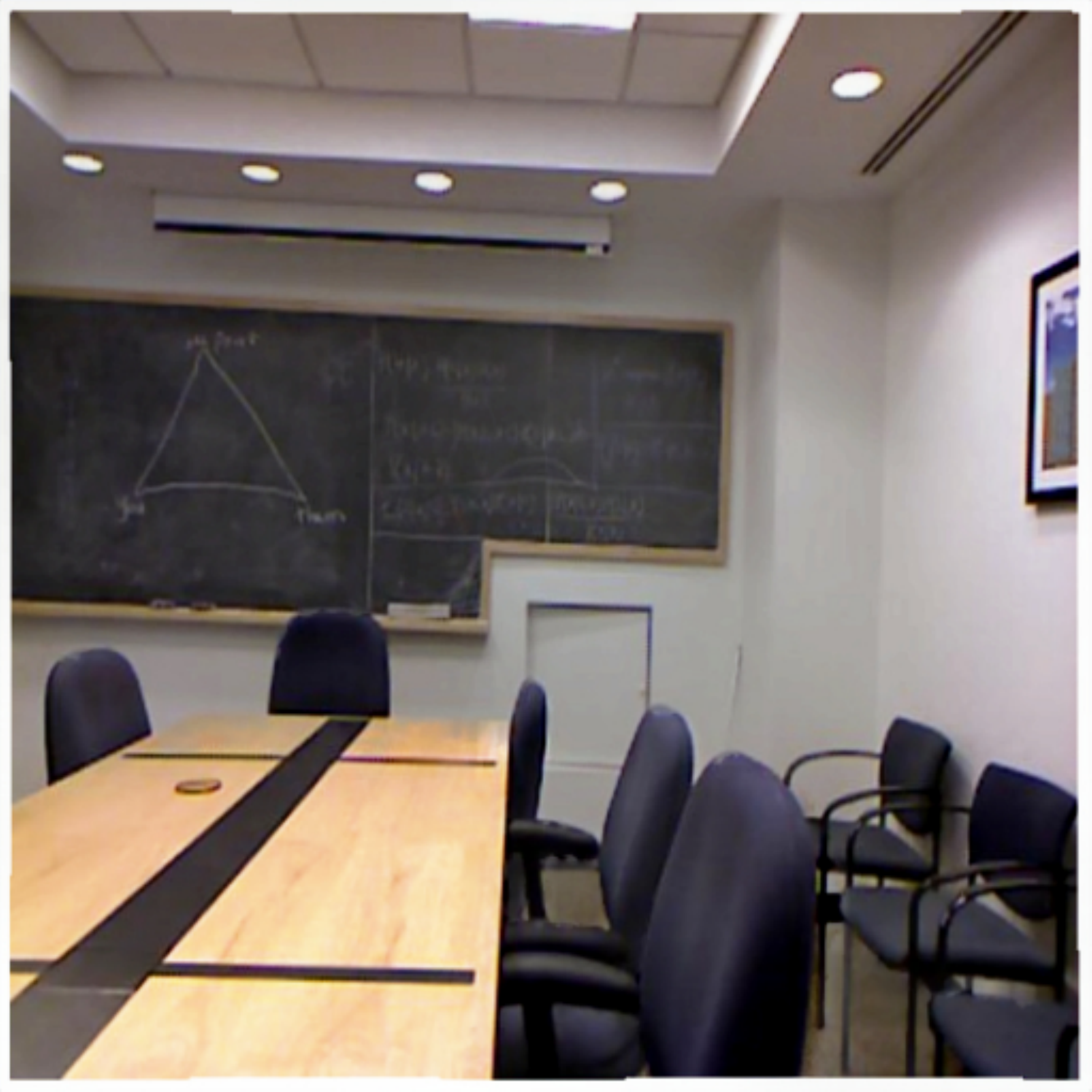}&
\includegraphics[width=0.14\linewidth]{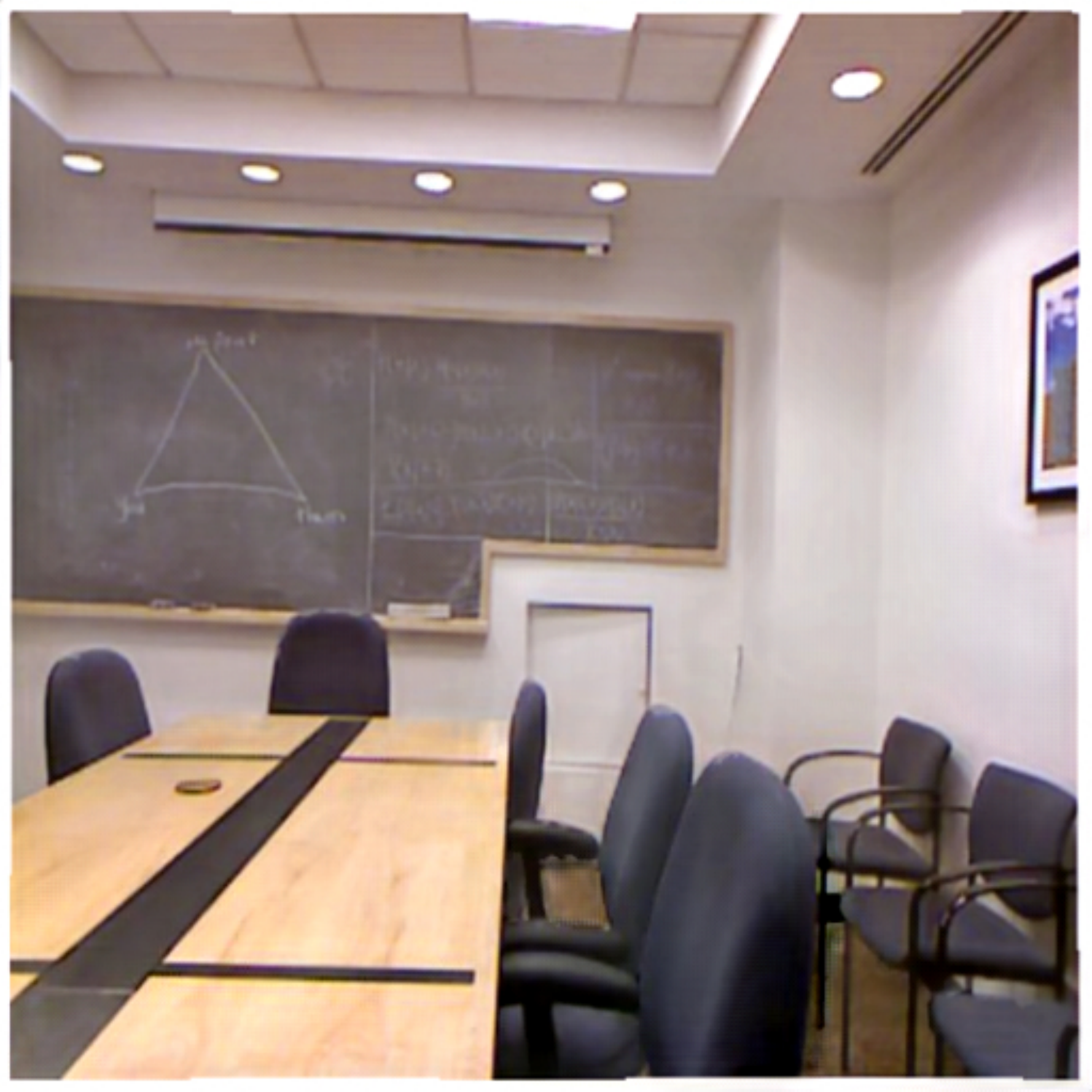}&
\includegraphics[width=0.14\linewidth]{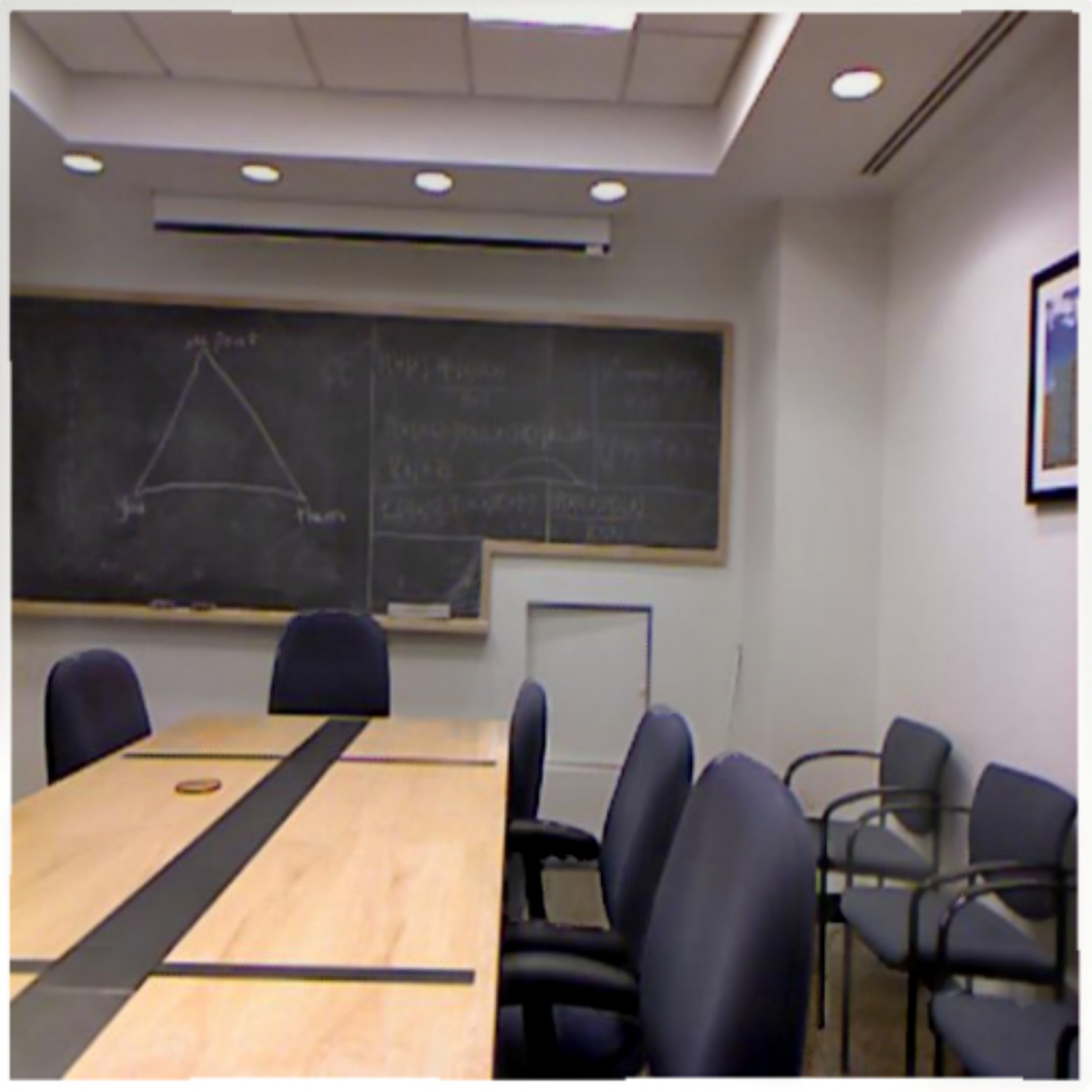} \\
 PSNR / SSIM & PSNR / SSIM &12.70 / 0.8068 & 15.29 / 0.8427 & 23.0591 / 0.9548 &  21.36 / 0.9457&  24.1392 / 0.9727\\
 Input & GT& DCP~\cite{he2010single} &AOD-Net~\cite{li2017aod} &DCPDN~\cite{zhang2018densely}  &GCANet~\cite{chen2018gated}  & PDLD (ours) \\
\end{tabular}
}
\caption{Dehazing comparisons for the indoor images.}
  \label{fig:indoor}
\end{figure*}
\begin{figure*}[ht]
\setlength{\tabcolsep}{1pt}
\centering
\footnotesize
\resizebox{1.0\linewidth}{!}
{
\centering
\begin{tabular}{ccc}
\includegraphics[width=0.33\linewidth]{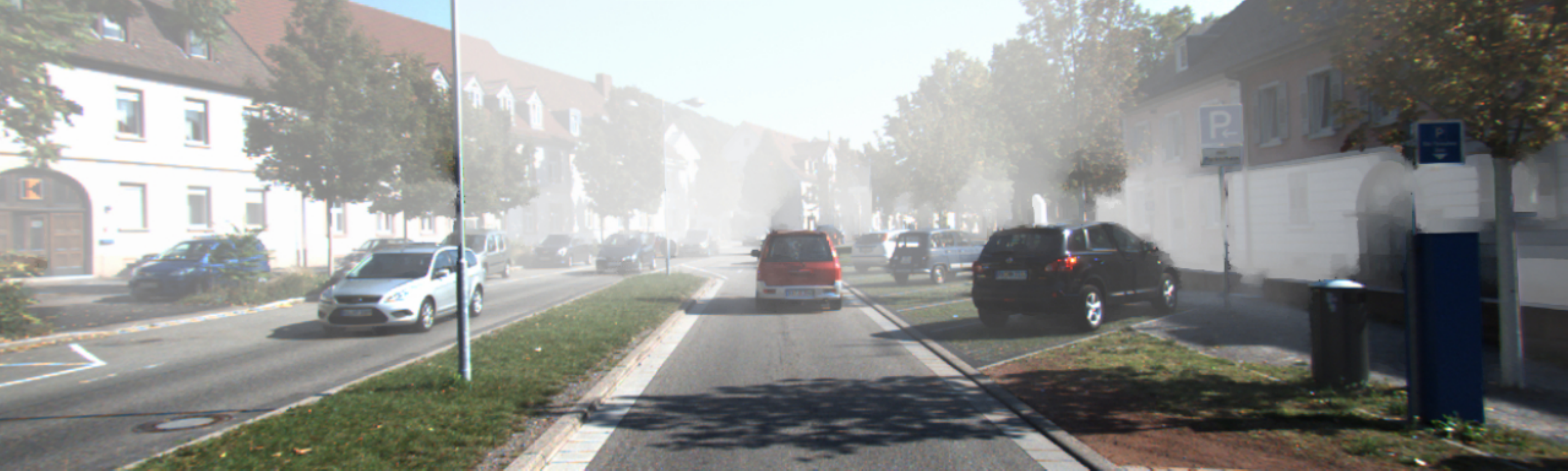}&
\includegraphics[width=0.33\linewidth]{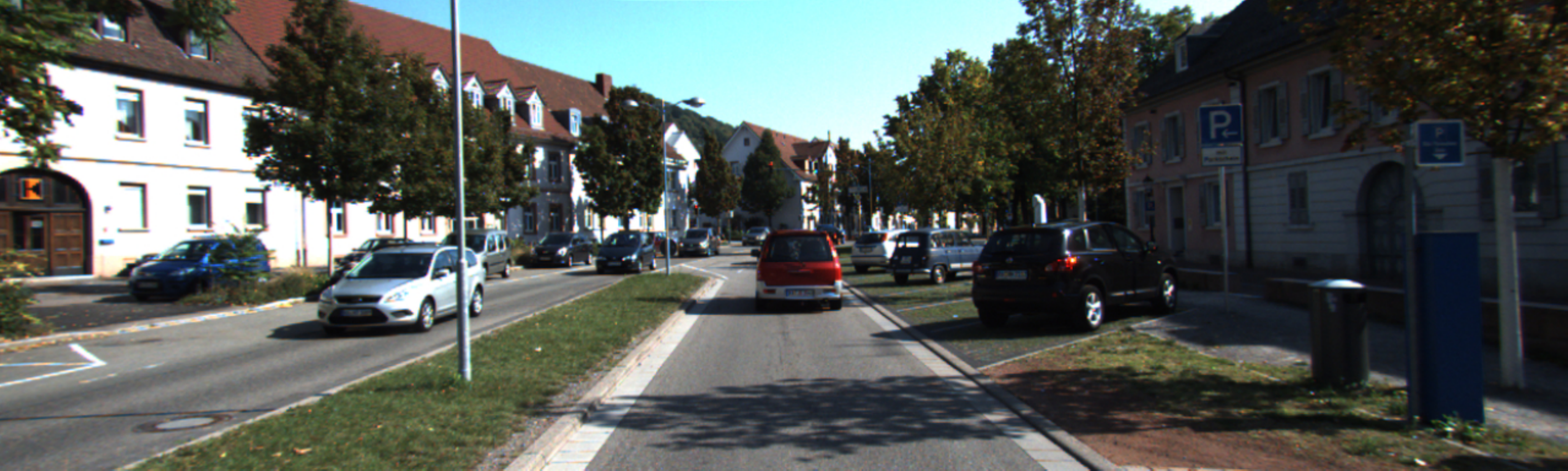}&
\includegraphics[width=0.33\linewidth]{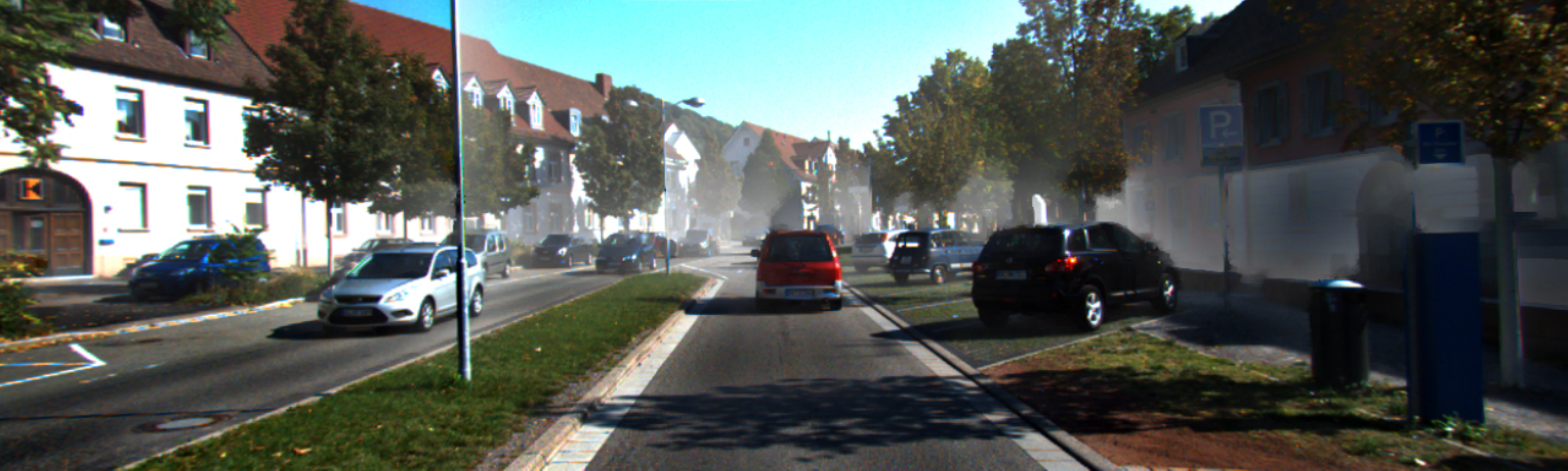}\\
PSNR / SSIM&  &  17.70 / 0.8666 \\
 Input & GT& DCP~\cite{he2010single} \\
\includegraphics[width=0.33\linewidth]{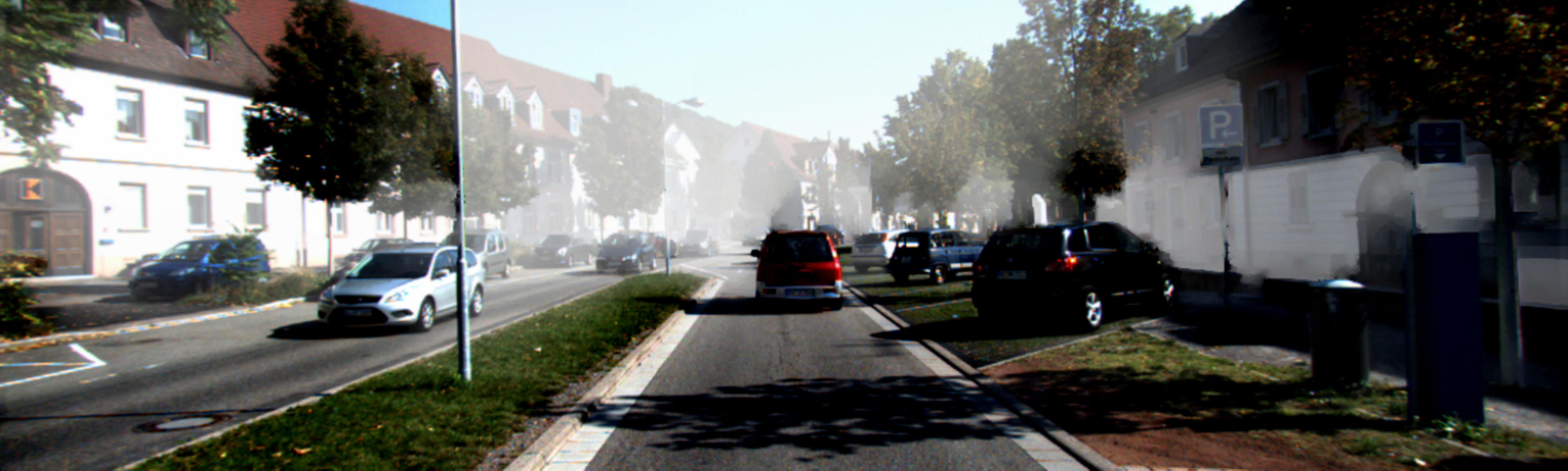}&
\includegraphics[width=0.33\linewidth]{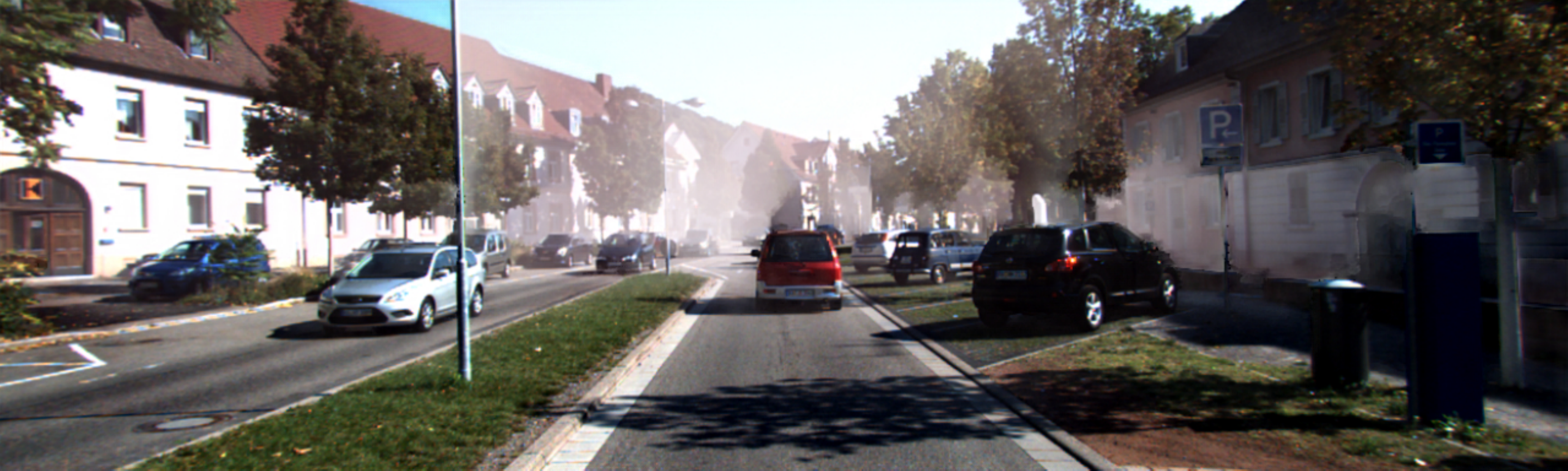}&
\includegraphics[width=0.33\linewidth]{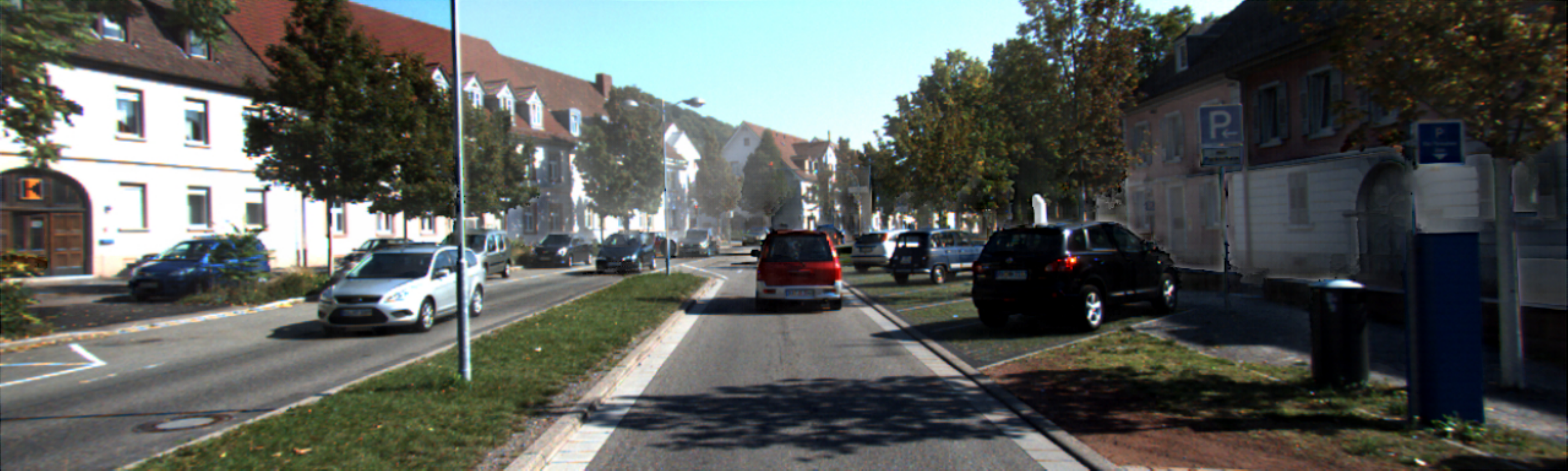}\\
  14.36 / 0.8001&  15.31 / 0.8332 & 22.66 / 0.9143\\
 DCPDN~\cite{zhang2018densely}  &GCANet~\cite{chen2018gated}  & PDLD(ours) \\

\includegraphics[width=0.33\linewidth]{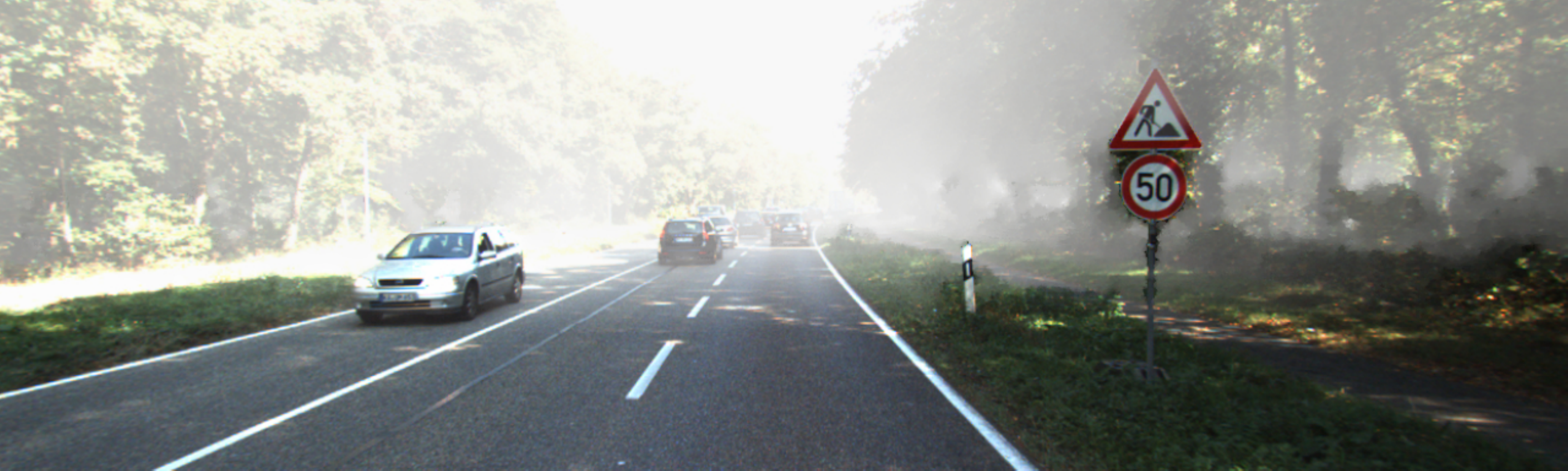}&
\includegraphics[width=0.33\linewidth]{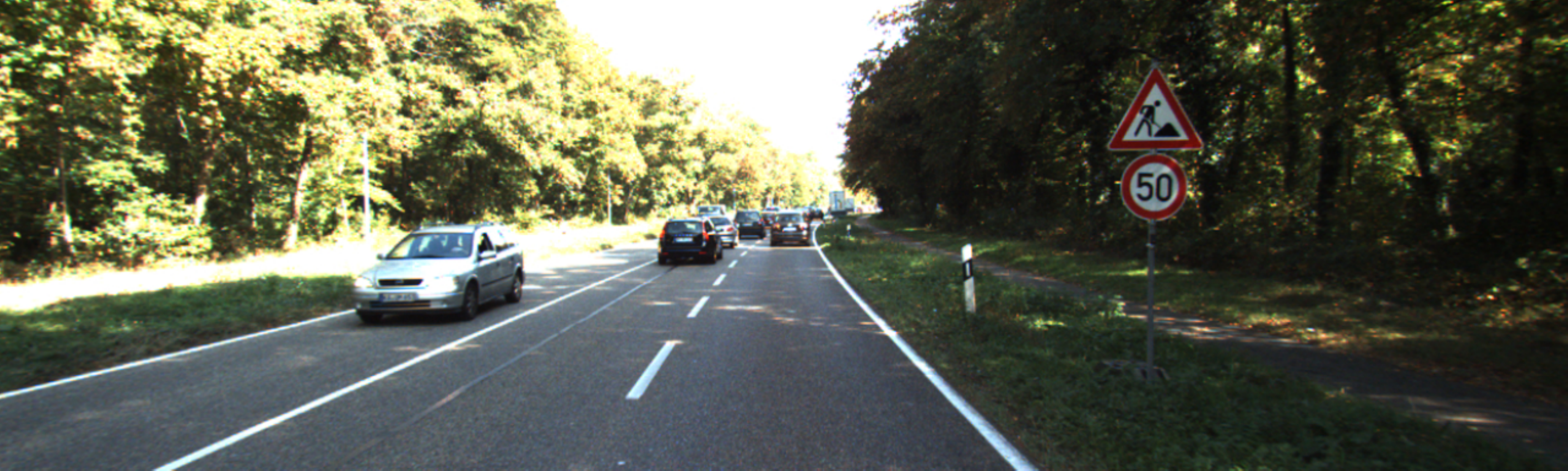}&
\includegraphics[width=0.33\linewidth]{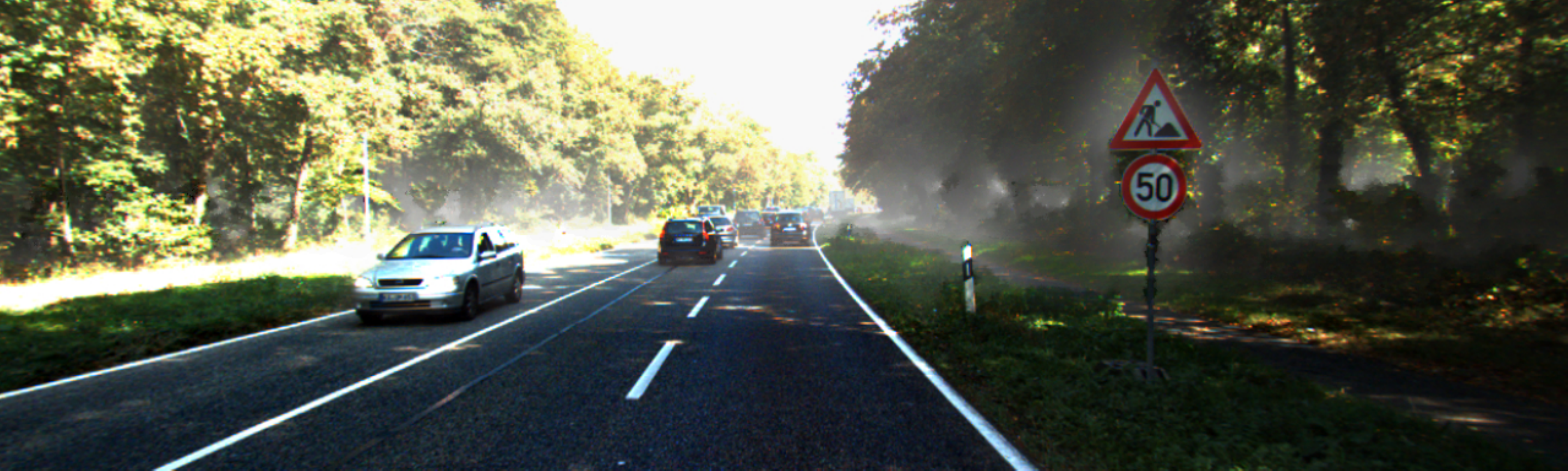}\\
PSNR / SSIM&  & 17.23 / 0.8151\\
 Input & GT& DCP~\cite{he2010single} \\
\includegraphics[width=0.33\linewidth]{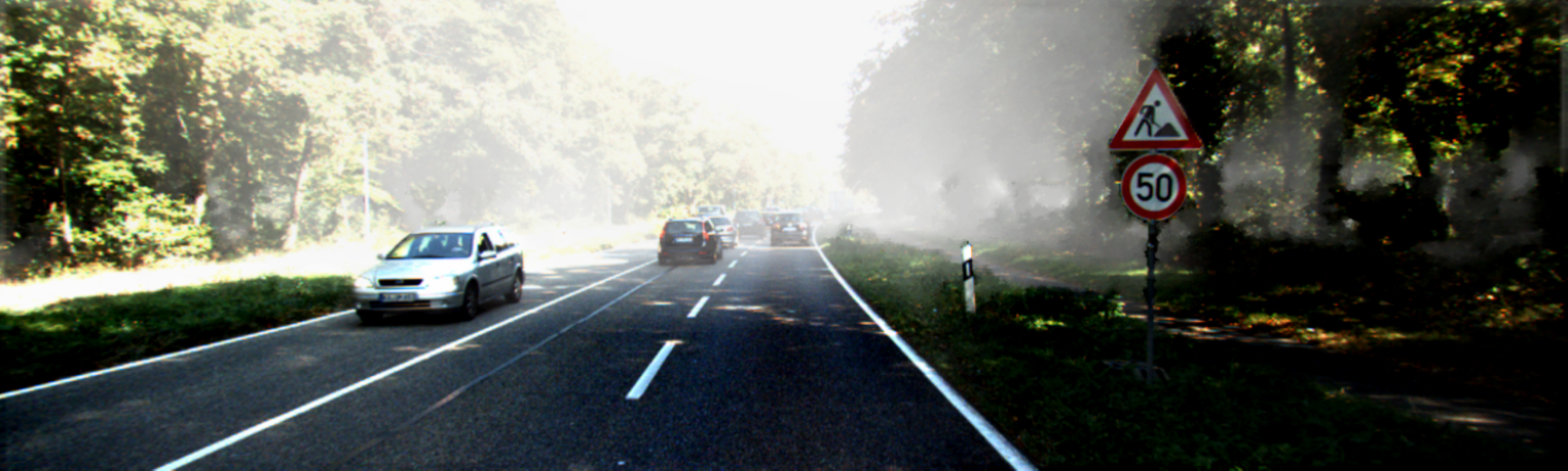}&
\includegraphics[width=0.33\linewidth]{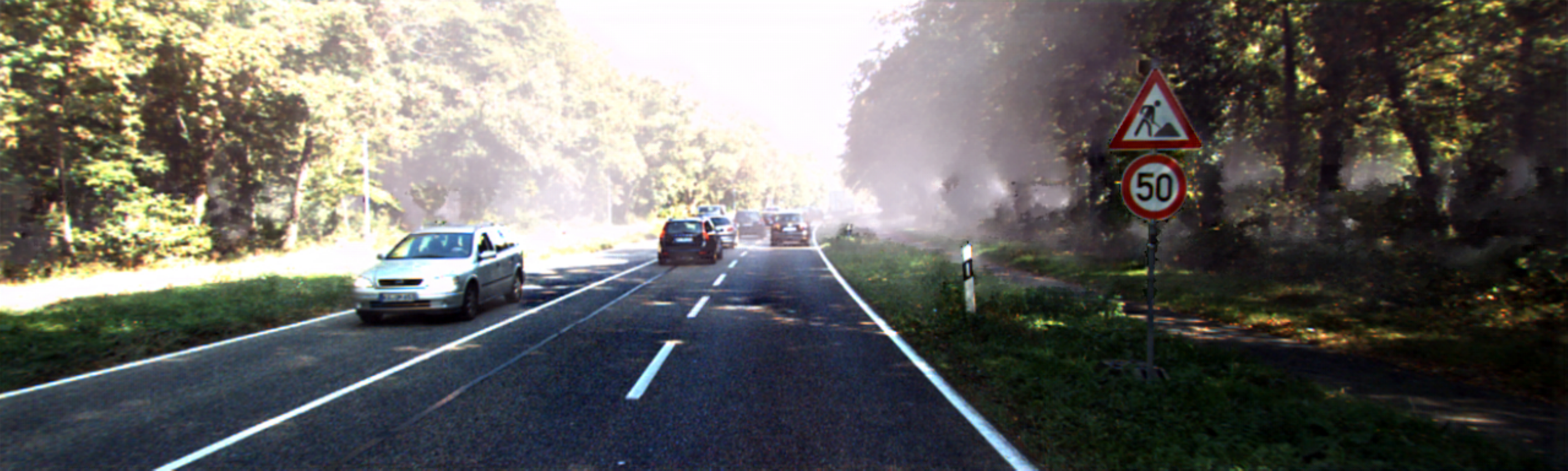}&
\includegraphics[width=0.33\linewidth]{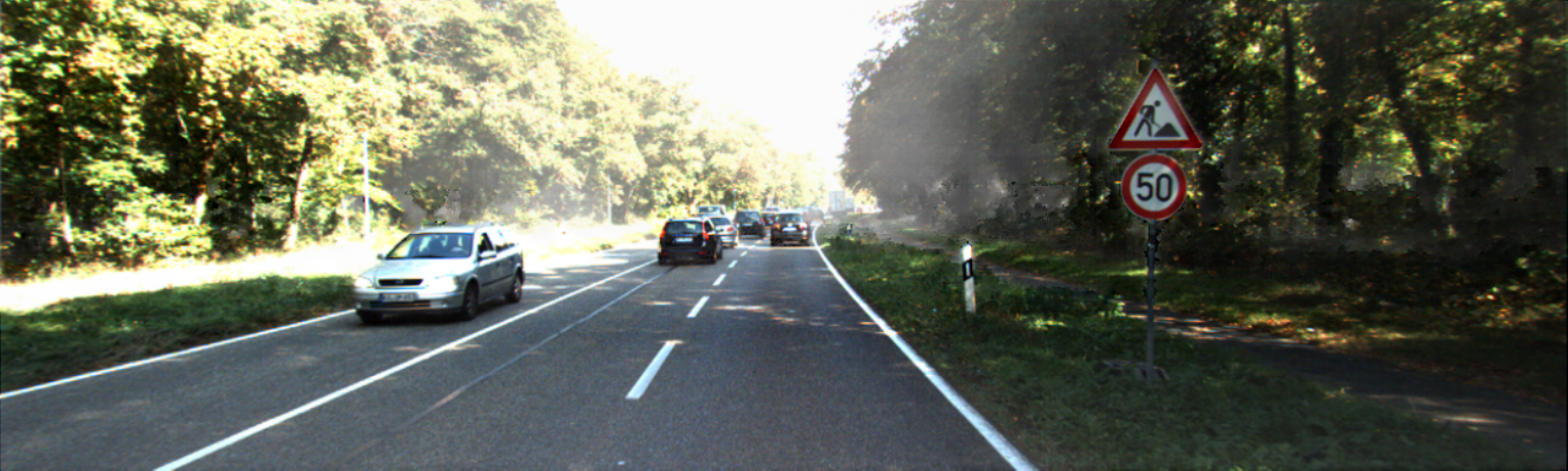}\\
  12.98 / 0.7735 &13.94 / 0.7806&19.7956 /0.9096 \\
 DCPDN~\cite{zhang2018densely}  &GCANet~\cite{chen2018gated}  & PDLD(ours) \\

\includegraphics[width=0.33\linewidth]{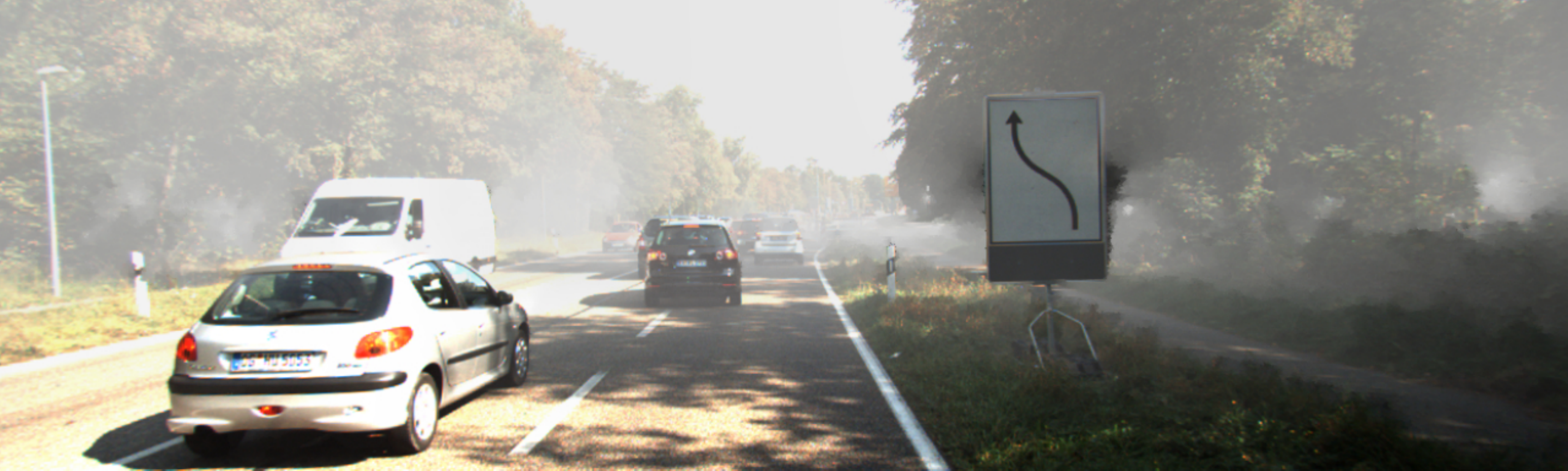}&
\includegraphics[width=0.33\linewidth]{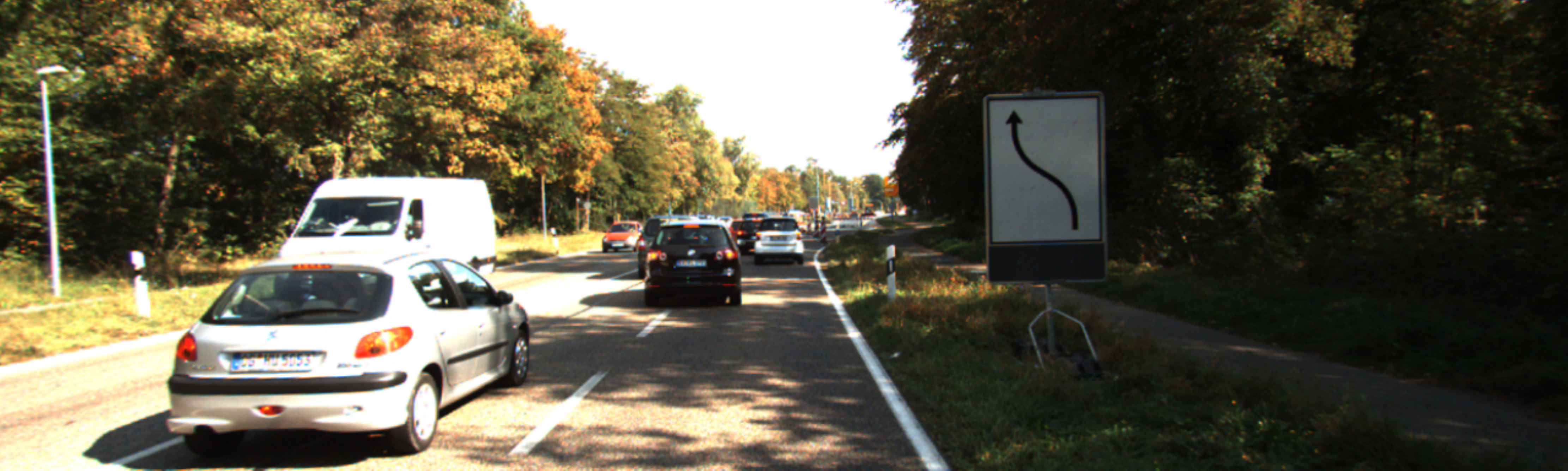}&
\includegraphics[width=0.33\linewidth]{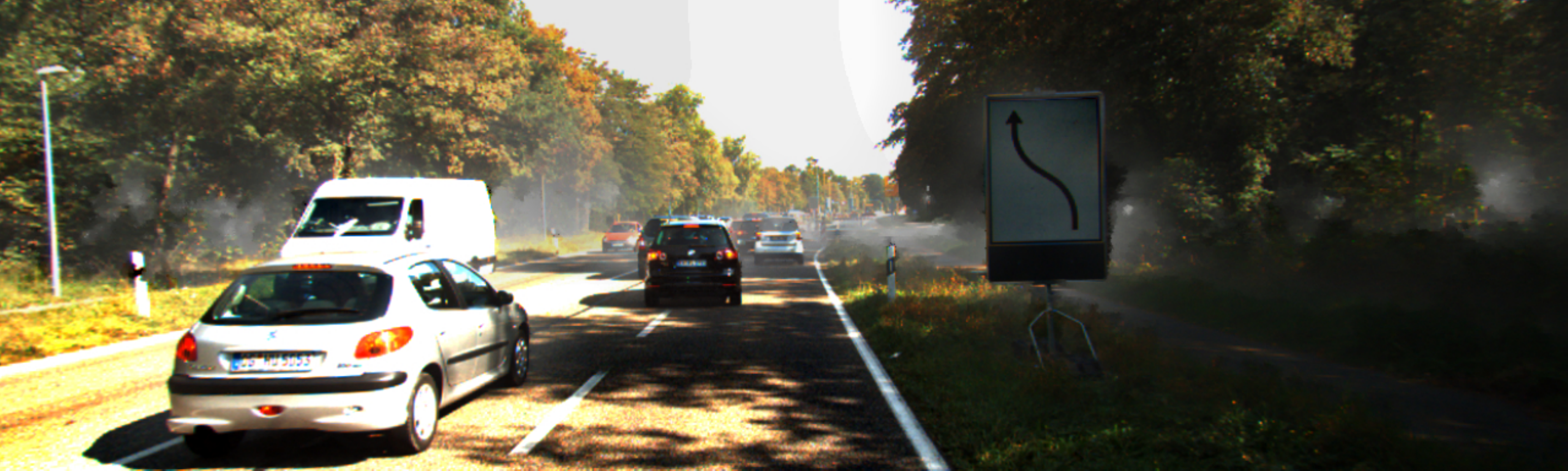}\\
PSNR / SSIM&  & 20.26 / 0.8653\\
 Input & GT& DCP~\cite{he2010single} \\
\includegraphics[width=0.33\linewidth]{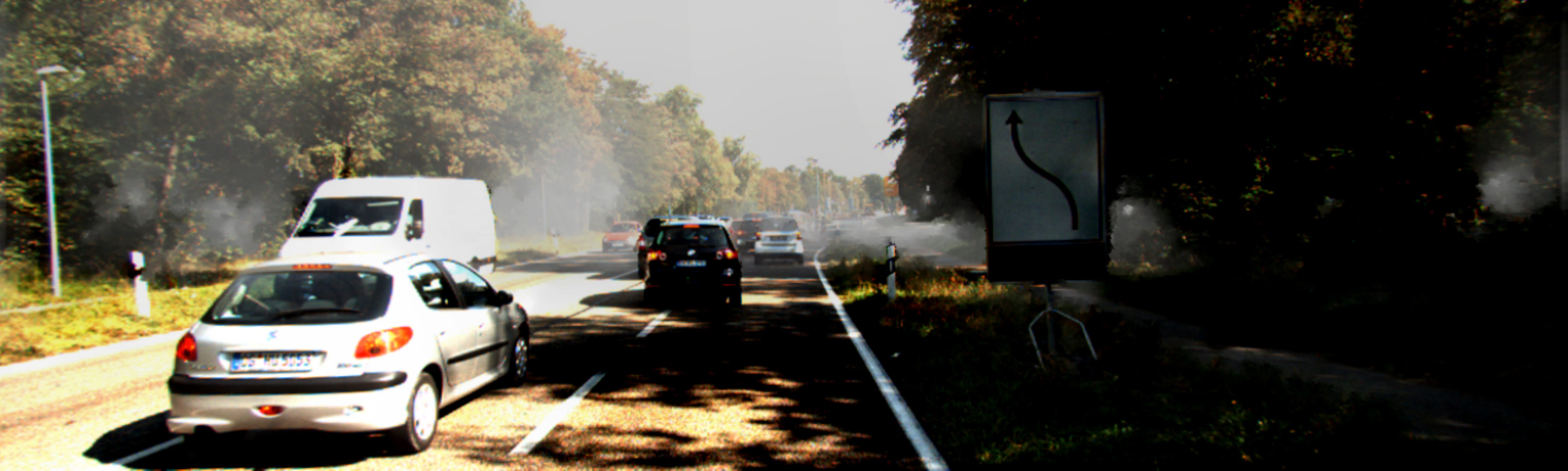}&
\includegraphics[width=0.33\linewidth]{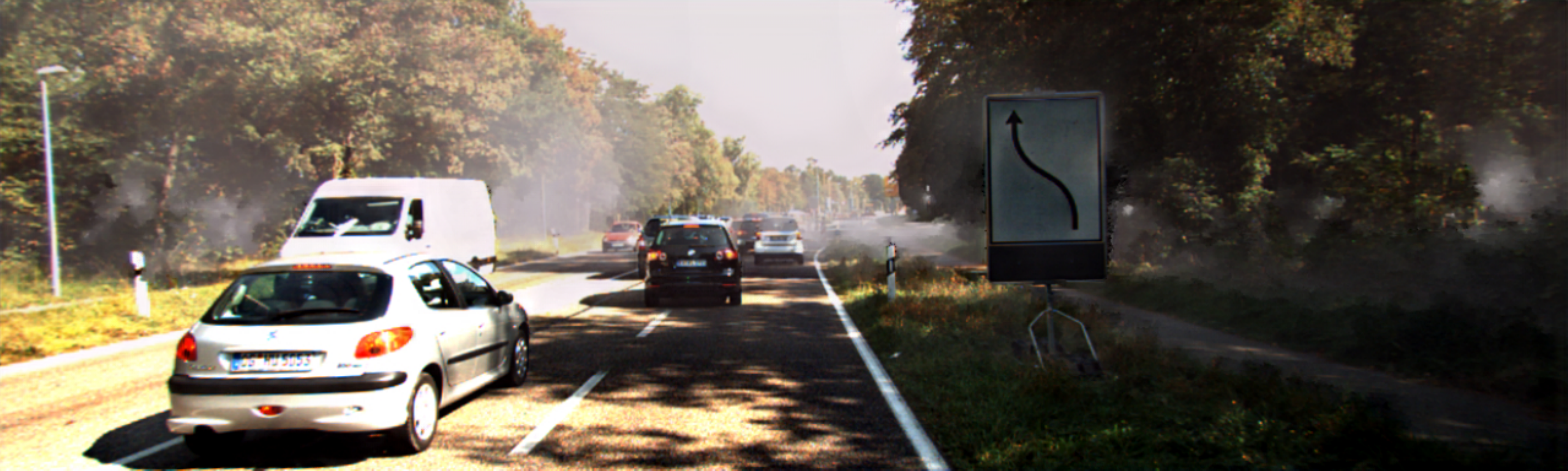}&
\includegraphics[width=0.33\linewidth]{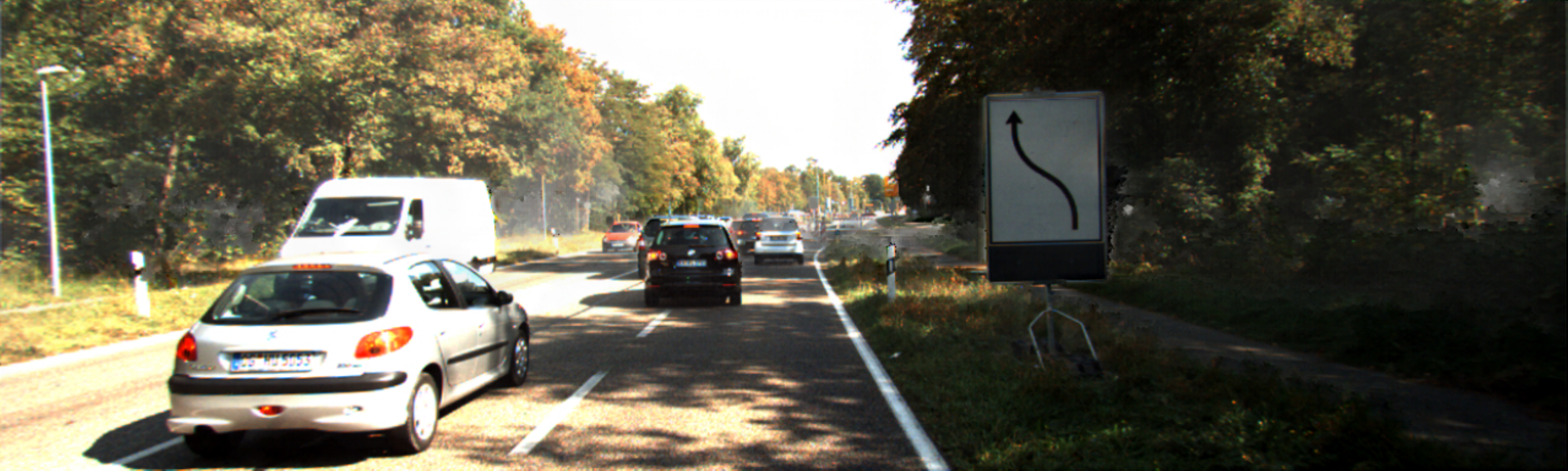}\\
 16.83 / 0.7876&17.70 / 0.8088&  23.59 / 0.9065\\
 DCPDN~\cite{zhang2018densely}  &GCANet~\cite{chen2018gated}  & PDLD(ours) \\
\end{tabular}
}
\caption{Dehazing comparisons visually for the outdoor images from KITTI2015.}
  \label{fig:outdoor}
\end{figure*}

Following \cite{ren2016single}, three atmospheric light conditions $A \in [0.7, 1.0]$ and the scattering coefficient $\beta \in [0.5, 1.5]$ are randomly sampled for each image to generate the corresponding hazy images. 6000 haze-free images are randomly selected from the training images of the NYU-depth2 dataset \cite{silberman2012indoor} to generate the 18000 hazy images as our training set \textbf{TrainA}. The images are resized to $512\times 512$, which is the same setting as DCPDN \cite{zhang2018densely} and the batchsize is set as 8.
Note that the compared method DCPDN \cite{zhang2018densely} applies four atmospheric light conditions where $A\in [0.7, 1.0]$ and $\beta\in [0.4, 1.6]$. For a fair comparison, we follow the \cite{zhang2018densely} to synthesize the testing set. The 654 images from the testing data of the NYU-depth2 dataset are applied to form $654*4$ testing images denoted as \textbf{TestA}. Although some test settings are not met in our training, our model performs excellently in the test set and outperforms all the compared methods in Test A. As shown in Fig.\ref{fig:indoor}, DCPDN and our methods are among the best to restore the dehazed images, while the AOD-Net fails to remove the hazes, DCP makes the images dark, and GCANet changes the brightness of the images. DCPDN fails in some regions such as distant places of the house in the first line of Fig.\ref{fig:indoor}.

\begin{table}
  \caption{Quantitative PSNR/SSIM comparisons on synthetic TestA.}
  \label{tab:testA}
  \scriptsize
  \setlength{\tabcolsep}{2pt}
  \begin{tabular}{cccccc}
    \toprule
                 & DCP   & AOD-Net &DCPDN  &GCANet  &PDLD\\
    \midrule
    Transmission & 15.18/0.8595& N/A          &25.92/0.9834 &N/A    &\textbf{26.90/0.9840}\\
    Dehazed      & 17.27/0.8587&17.72/0.8577&28.00/0.9496 &25.88/0.9211 &\textbf{28.23/0.9523} \\
    \bottomrule
  \end{tabular}
\end{table}

To demonstrate the generalization ability of our indoor model, the model trained by the NYU-depth2 dataset is applied on the Middleburry stereo dataset~\cite{scharstein2014high}.
23 images with ground truths from the Middlebury stereo dataset are synthesized to form $23*4$ images as \textbf{TestB} following \cite{zhang2018densely}. The quantitative comparisons on the TestB are reported in Table \ref{tab:Middleury}, which demonstrates that our model exhibits excellent generalization abilities and achieves favorable results on the Middlebury stereo dataset.

\begin{table}
  \caption{Quantitative PSNR/SSIM comparisons on synthetic TestB.}
  \label{tab:Middleury}
  \scriptsize
  \setlength{\tabcolsep}{3pt}
  \begin{tabular}{cccccc}
    \toprule
                 &DCP&AOD-Net&DCPDN&GCANet&PDLD\\
    \midrule
    Transmission &10.93/0.6836& N/A  &16.17/0.8521&N/A&\textbf{16.31/0.8714} \\
    Dehazed      &16.28/0.8413&17.8037/0.8449&20.04/0.8801&\textbf{20.79/0.9003}&20.41/0.8973\\
    \bottomrule
  \end{tabular}
\end{table}

\subsection{Outdoor Dataset}

\textbf{KITTI} \cite{Geiger2013IJRR} and \textbf{KITTI 2015} \cite{Menze2015CVPR} are real-world datasets of street views from an autonomous driving platform, and sparse ground-truth depth information is obtained with LiDAR. For the real outdoor dataset, the performance of both dehazing and depth estimations is provided to demonstrate the effectiveness of our progressive depth estimation and image dehazing strategy. The depth estimation has largely improved the transmission estimations and dehazed results of the outdoor dataset.

To train the proposed network, 6000 images are randomly selected from the training set of the KITTI raw data.  In addition, 697 test images of the KITTI raw data and KITTI 2015 are used to compare our algorithm against the state-of-the-art method~\cite{song2020simultaneous}. Images are resized to $1280 \times 384$ resolution for the training and testing. Although depth estimation by stereo matching is easier than single image depth estimation, it is surprising that our method outperforms \cite{song2020simultaneous} in the depth estimations and dehazing tasks.


\begin{table}
  \caption{Quantitative PSNR / SSIM comparisons on KITTI2015 dataset.}
  \label{tab:KITTI}
  \scriptsize
  \begin{tabular}{ccccccc}
    \toprule
    & DCP   & AOD-Net & DCPDN  & GCANet  & PDLD   &~\cite{song2020simultaneous}\\
    \midrule
     Transmission(PSNR) & 14.82&	N/A & 15.27	N/A &\textbf{25.57}& N/A\\
    Transmission(SSIM) & 0.8385& N/A     &0.8813 &N/A    &\textbf{0.9675} & N/A\\
     Dehazed(PSNR)& 17.19	&13.38	&15.67	&16.86	&24.94 &23.02\\
    Dehazed(SSIM)      & 0.8431&0.7357   & 0.8054 &0.8352 &0.9275         &0.853\\
    \bottomrule
  \end{tabular}
\end{table}

\textbf{FRIDA 3}~\cite{jpt-cva14} contains 66 foggy outdoor road scenes, which is synthetically generated clear/hazy pairs of stereo images (Caraffa and Tarel 2012).  Note that \cite{song2020simultaneous} adapted the model on the FRIDA 3 dataset, our approach adapted with the same settings is reported for FRIDA 3 which achieves the best performance.

\begin{table}
  \caption{Quantitative PSNR / SSIM comparisons of the dehazed images on FRIDA 3 dataset.}
  \label{tab:FRIDA3}
  \scriptsize
  \begin{tabular}{cccccccc}
    \toprule
                 & DCP   & AOD-Net &DCPDN  &GCANet  &PDLD   & PDLD adapted&  ~\cite{song2020simultaneous}\\
    \midrule
    PSNR     & 12.44&10.44 & 10.88 &12.80 &13.73&15.27 &\textbf{16.59}\\
    SSIM     & 0.7151& 0.7925 & 0.7375 &0.7126 &0.7958&\textbf{0.8737} &0.8387\\
    \bottomrule
  \end{tabular}
\end{table}

\textbf{Cityscapse} is also tested for a random selected 200 images in table.\ref{tab:FRIDA3}.
\begin{table}
 \caption{Quantitative PSNR / SSIM comparisons on Cityscapse dataset.}
  \label{tab:FRIDA3}
    \scriptsize
  \begin{tabular}{ccccccc}
    \toprule
                 & DCP   & AOD-Net &DCPDN  &GCANet  &PDLD   &  ~\cite{song2020simultaneous}\\
    \midrule
      Transmission(PSNR) & 18.93	&N/A 	&16.27	&N/A 	&	\textbf{20.36}& N/A\\
    Transmission(SSIM) & 0.9047& N/A     &0.8620 &N/A    &\textbf{0.9675} & N/A\\
     Dehazed(PSNR)      &15.64	&14.88	&13.79	&16.23 &	 \textbf{22.14}&\textbf{23.72}\\
     Dehazed(SSIM)      &0.7335&0.7152   & 0.7974 &0.7733 & \textbf{0.9120} & 0.876\\

    \bottomrule
  \end{tabular}
\end{table}

\begin{figure*}[ht]
\setlength{\tabcolsep}{1pt}
\centering
\footnotesize
\resizebox{1.0\linewidth}{!}
{
\centering
\begin{tabular}{ccc}


\includegraphics[width=0.33\linewidth]{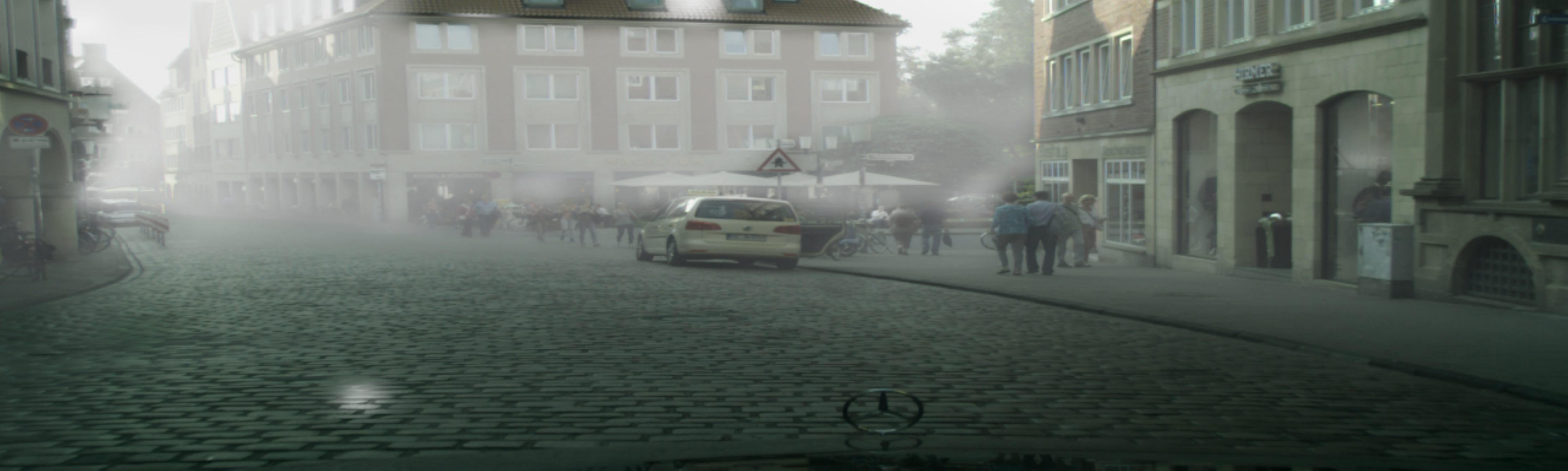}&
\includegraphics[width=0.33\linewidth]{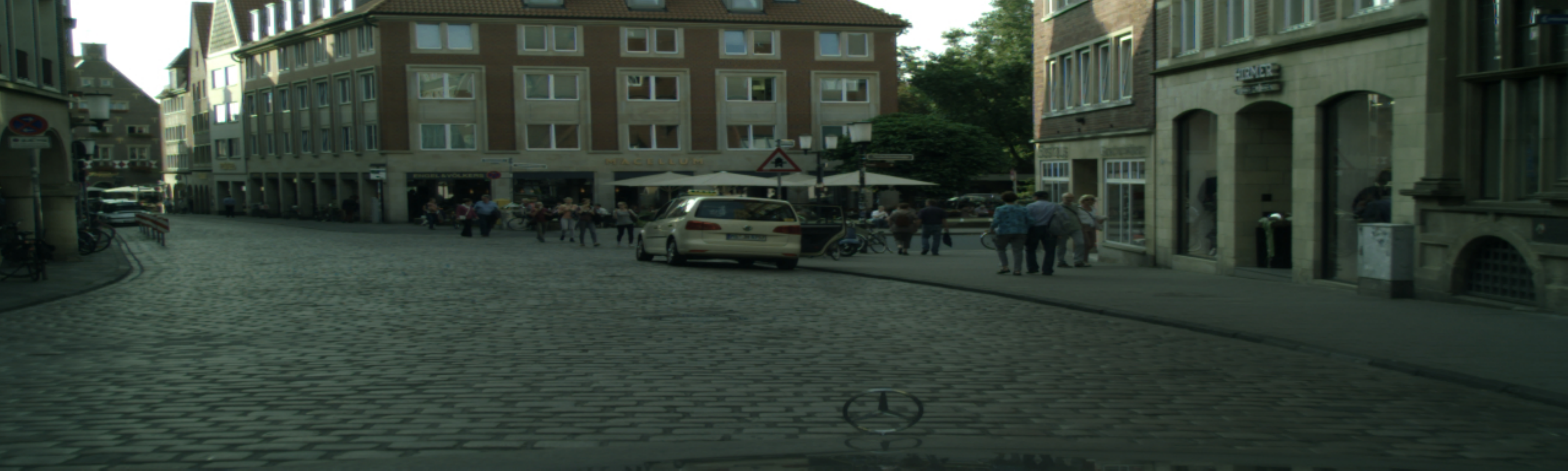}&
\includegraphics[width=0.33\linewidth]{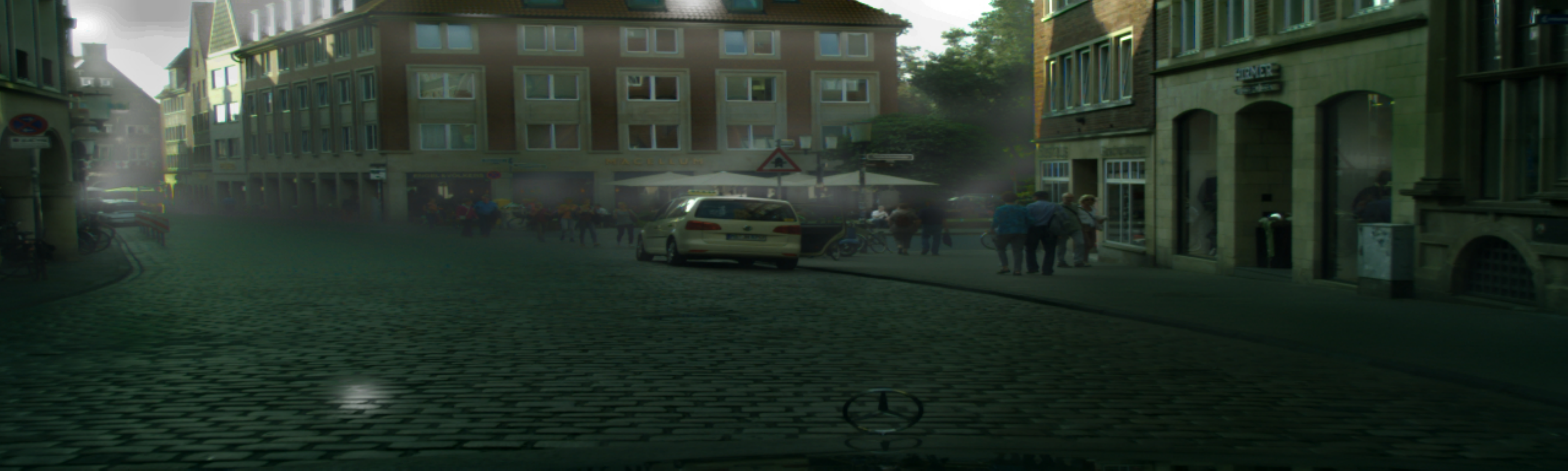}\\
PSNR / SSIM& &20.13/0.8288\\
Input & GT& DCP~\cite{he2010single} \\
\includegraphics[width=0.33\linewidth]{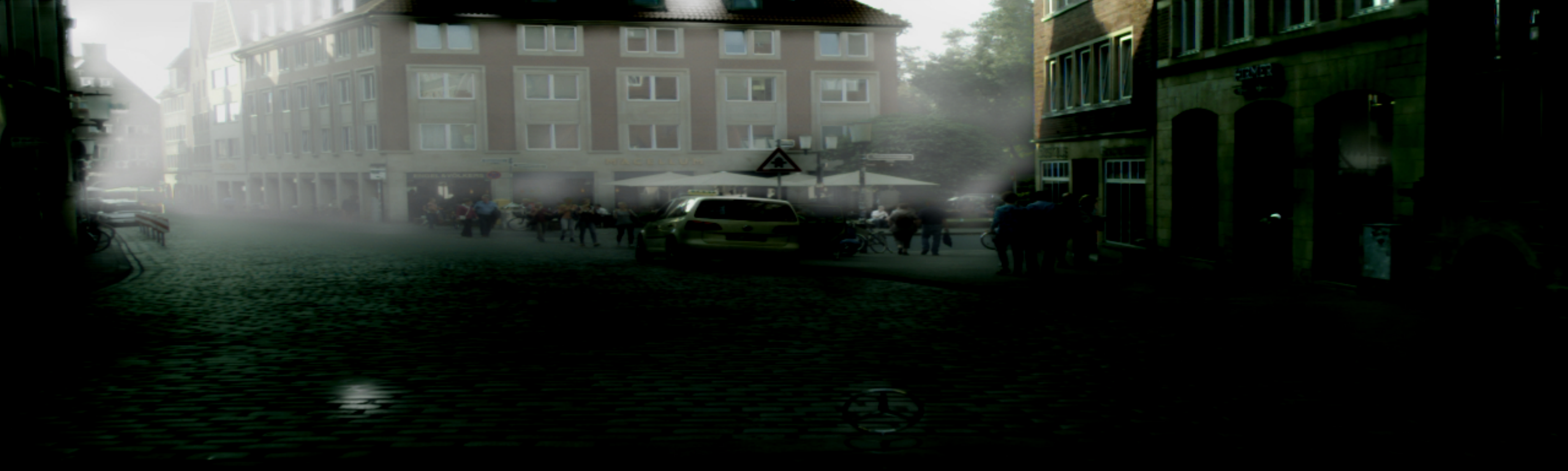}&
\includegraphics[width=0.33\linewidth]{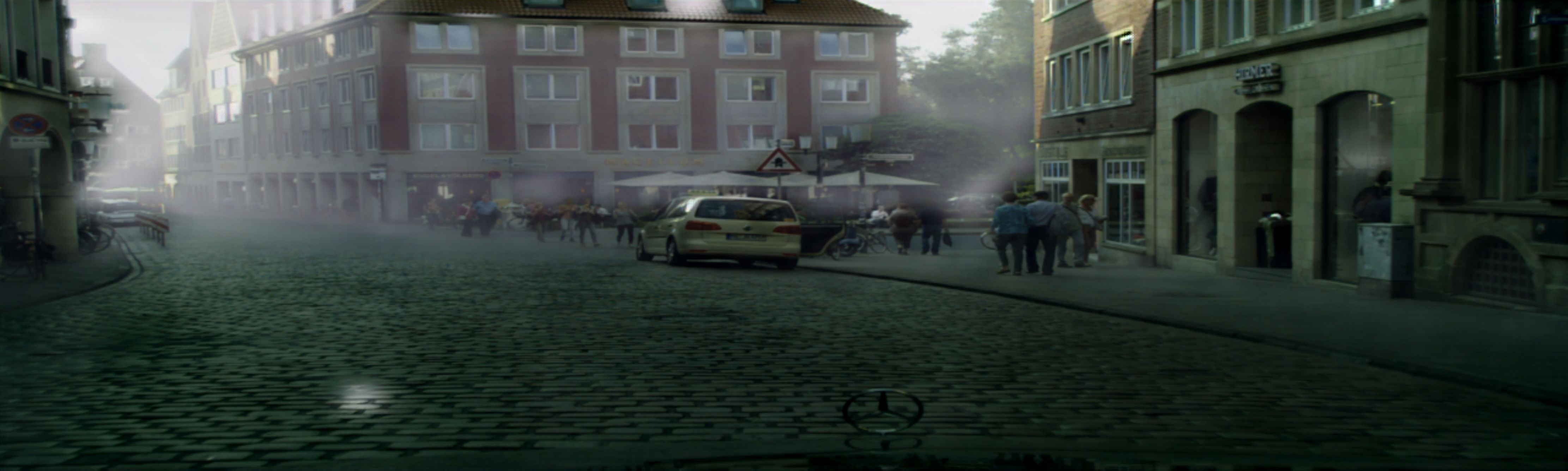}&
\includegraphics[width=0.33\linewidth]{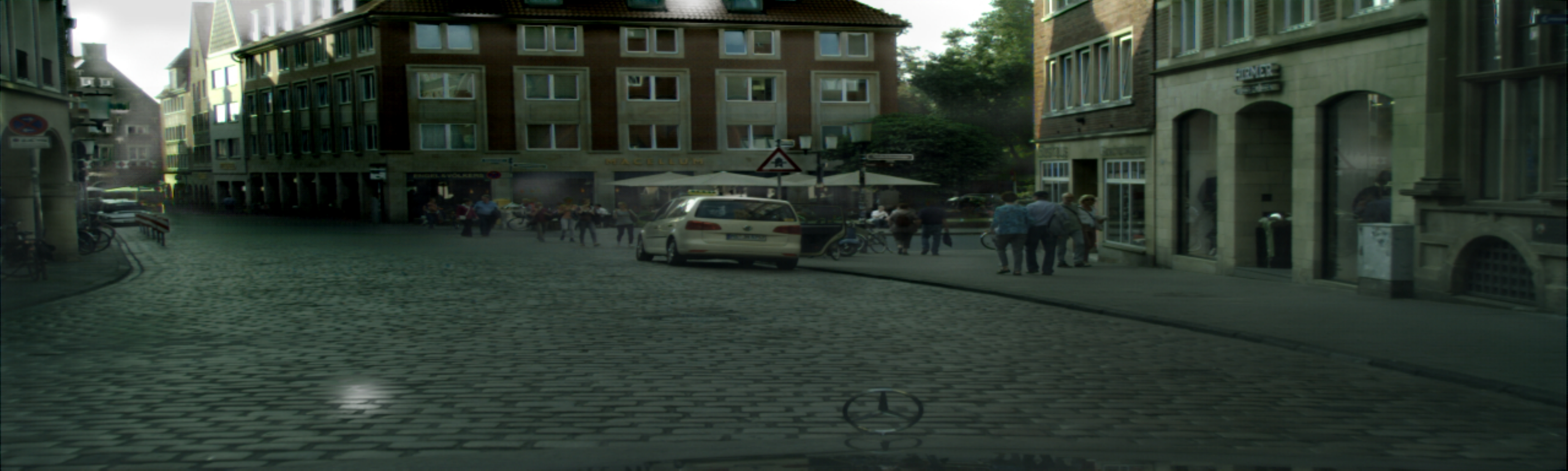}\\
   14.18 / 0.7785 &  17.15 / 0.8054 &   22.40 / 0.9130\\
 DCPDN~\cite{zhang2018densely}&GCANet~\cite{chen2018gated}  & PDLD(ours) \\

\includegraphics[width=0.33\linewidth]{cityscape/127haze}&
\includegraphics[width=0.33\linewidth]{cityscape/127gt}&
\includegraphics[width=0.33\linewidth]{cityscape/127dcpdehaze}\\
PSNR / SSIM&  &  11.16 / 0.6508\\
Input & GT& DCP~\cite{he2010single} \\
\includegraphics[width=0.33\linewidth]{cityscape/127dcpdndehaze}&
\includegraphics[width=0.33\linewidth]{cityscape/127gcadehaze}&
\includegraphics[width=0.33\linewidth]{cityscape/127pdlddehaze}\\
13.20 / 0.7134& 15.85 / 0.7276 & 20.92 / 0.9133\\
 DCPDN~\cite{zhang2018densely} &GCANet~\cite{chen2018gated}  & PDLD(ours) \\
\end{tabular}
}
\caption{Dehazing comparisons visually for the outdoor images from Cityscapse.}
  \label{fig:outdoor2}
\end{figure*}

Our method achieves the best objective evaluation performances in vast majorities of the experiments and performs especially well in the outdoor dataset. From Fig.\ref{fig:outdoor} and Fig.\ref{fig:outdoor2}, compared with the groundtruth images, our methods has a great advantage for the distant areas with the thick hazes, since depth estimations has constrained the transmission estimation process. Almost all the compared methods either change the brightness or leave thick hazes.Our approach preserves the details and original brightness as well as remove the heavy hazes. Outdoor hazy images have larger distances and thicker hazes for which predicting transmissions are more challenging problem. Progressive depth learning and transmission estimation lead to better depth estimation and dehazing performances.



\begin{figure*}[ht]
\setlength{\tabcolsep}{1pt}
\centering
\footnotesize
\resizebox{1.0\linewidth}{!}
{
\centering
\begin{tabular}{cccccc}

\includegraphics[width=0.16\linewidth]{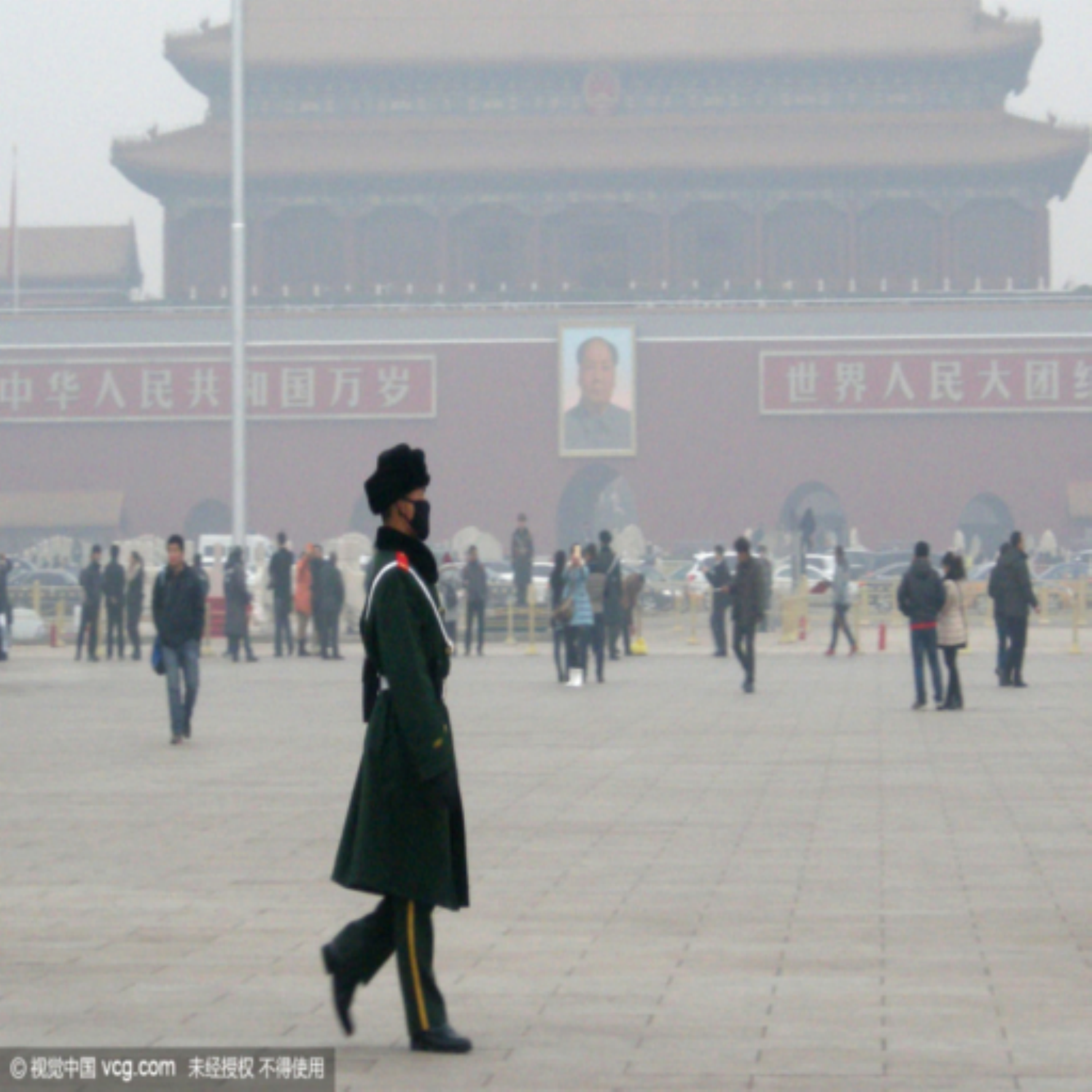}&
\includegraphics[width=0.16\linewidth]{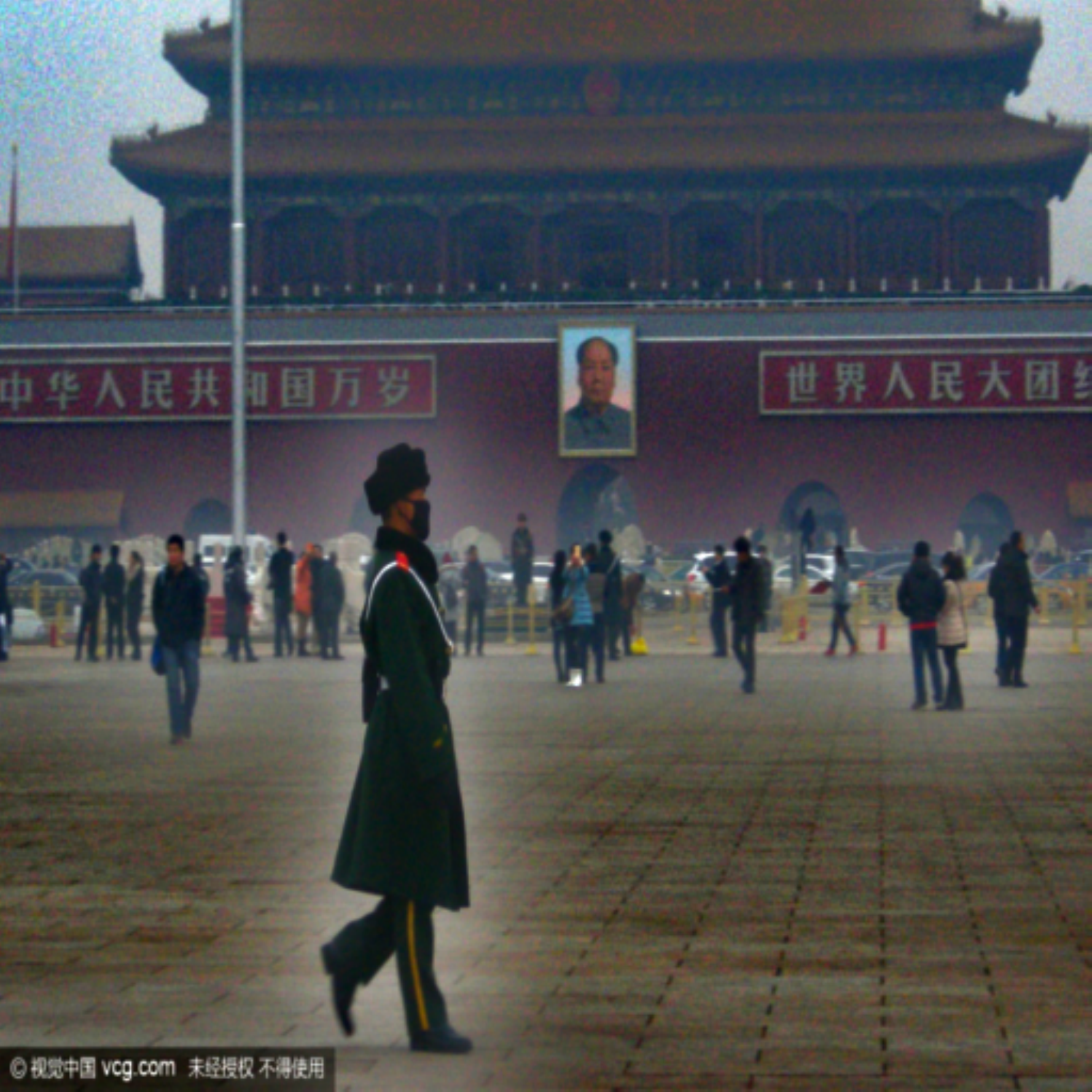}&
\includegraphics[width=0.16\linewidth]{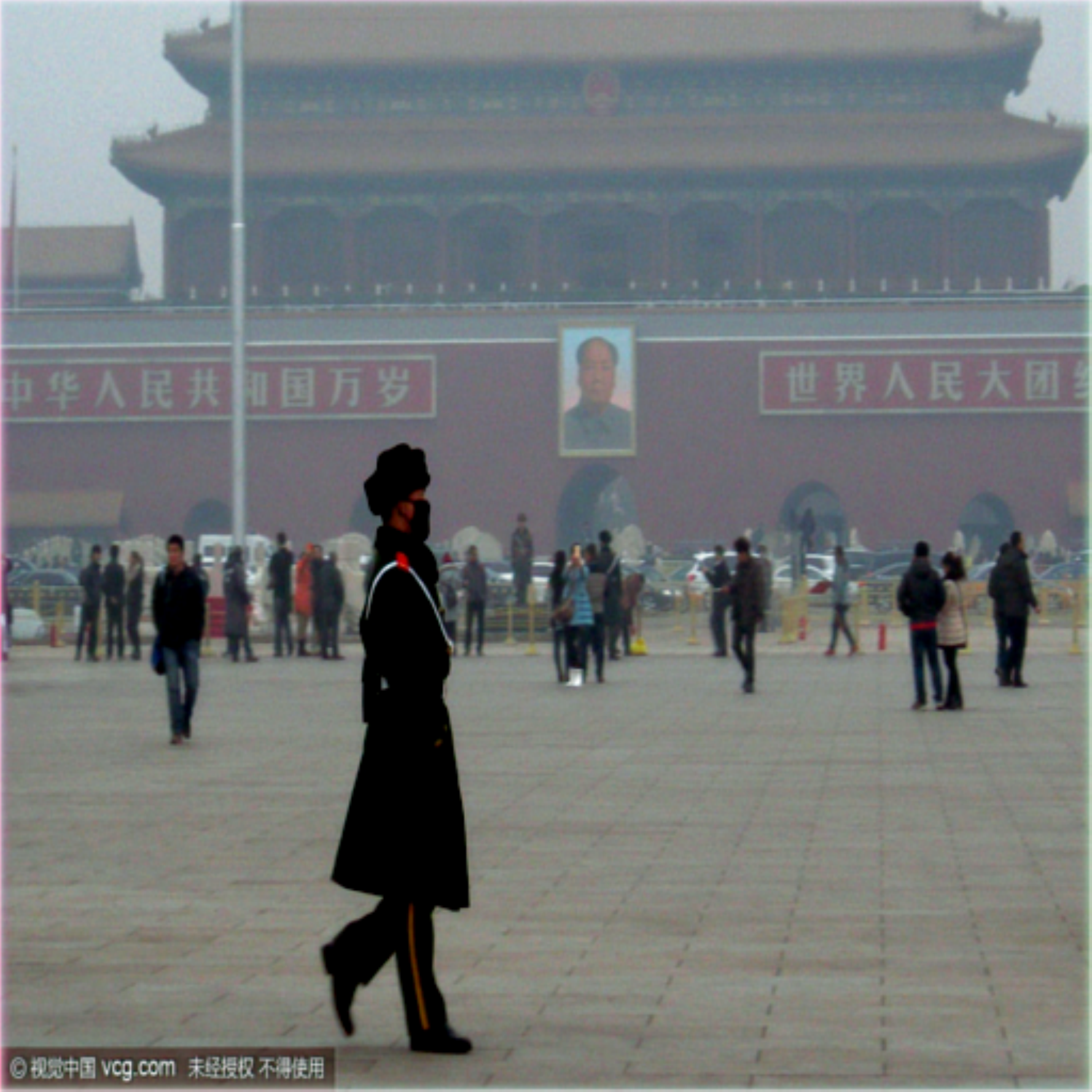}&
\includegraphics[width=0.16\linewidth]{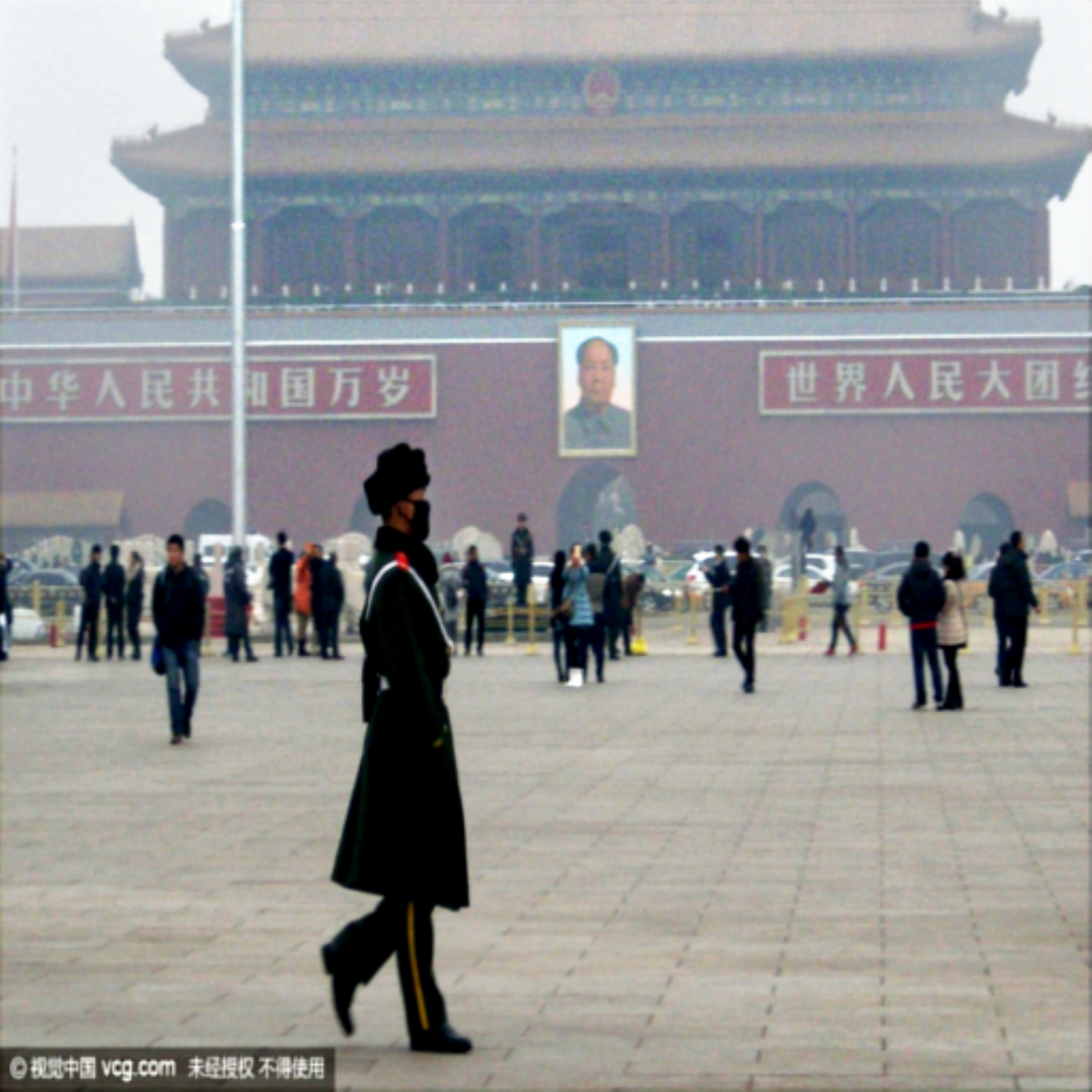}&
\includegraphics[width=0.16\linewidth]{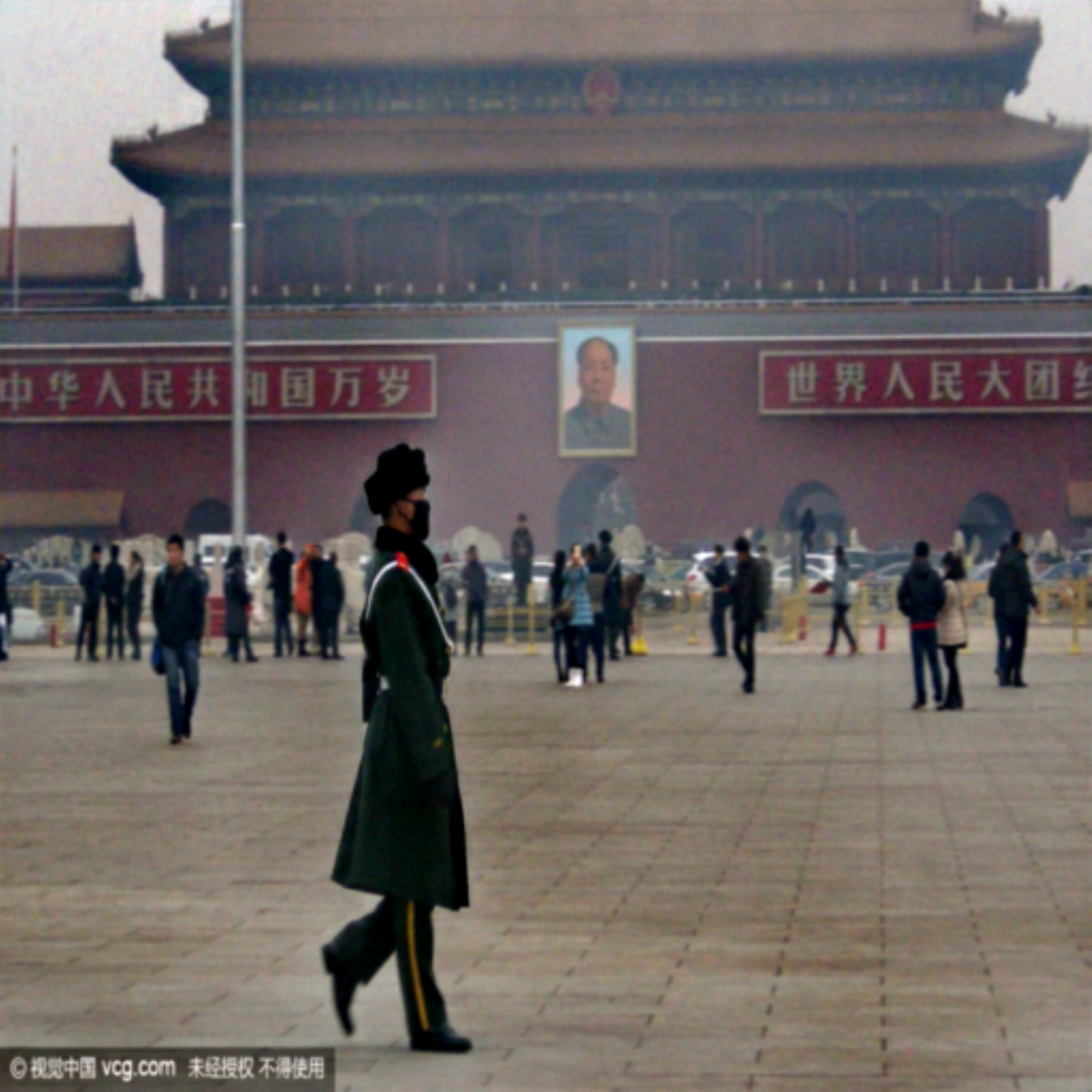}&
\includegraphics[width=0.16\linewidth]{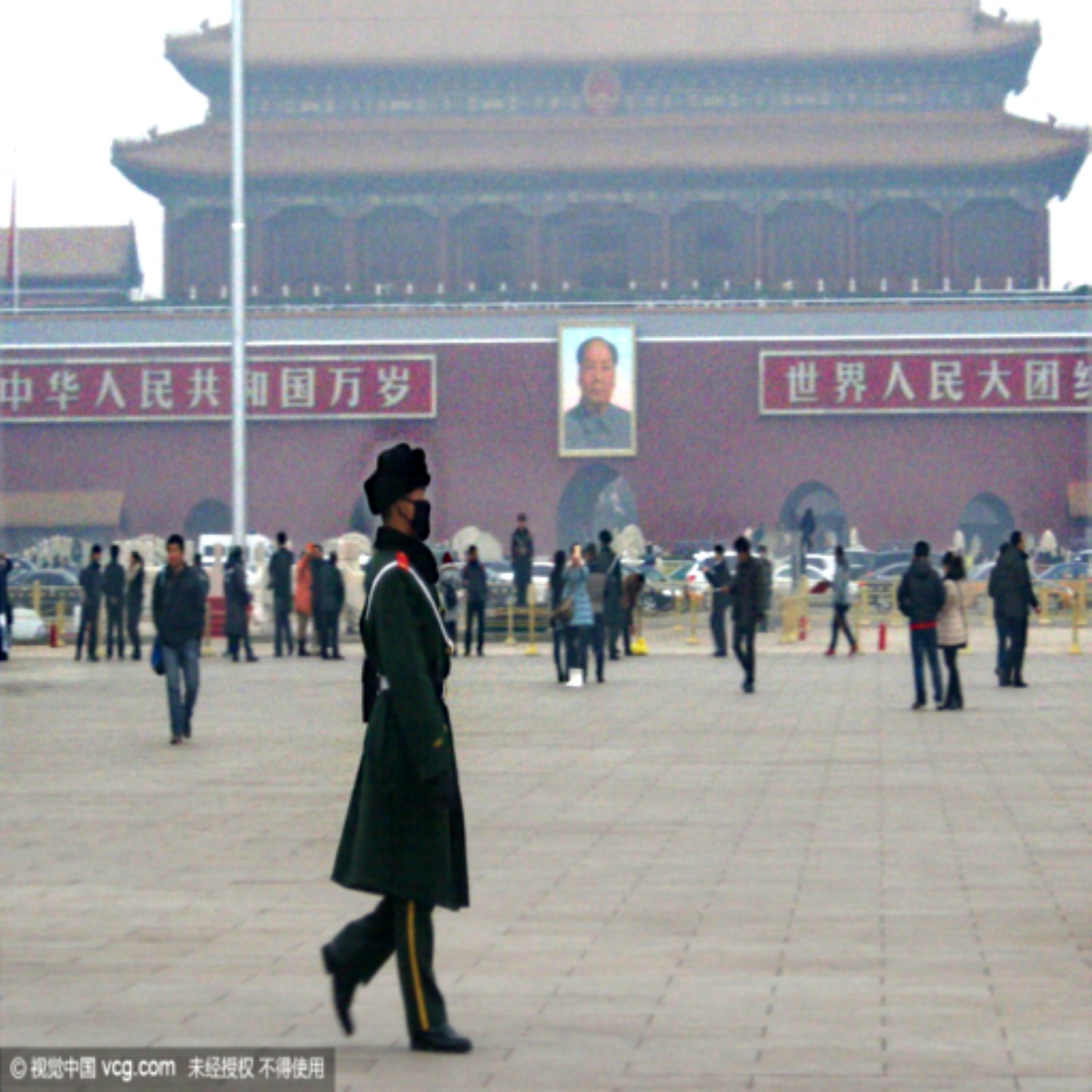}\\

\includegraphics[width=0.16\linewidth]{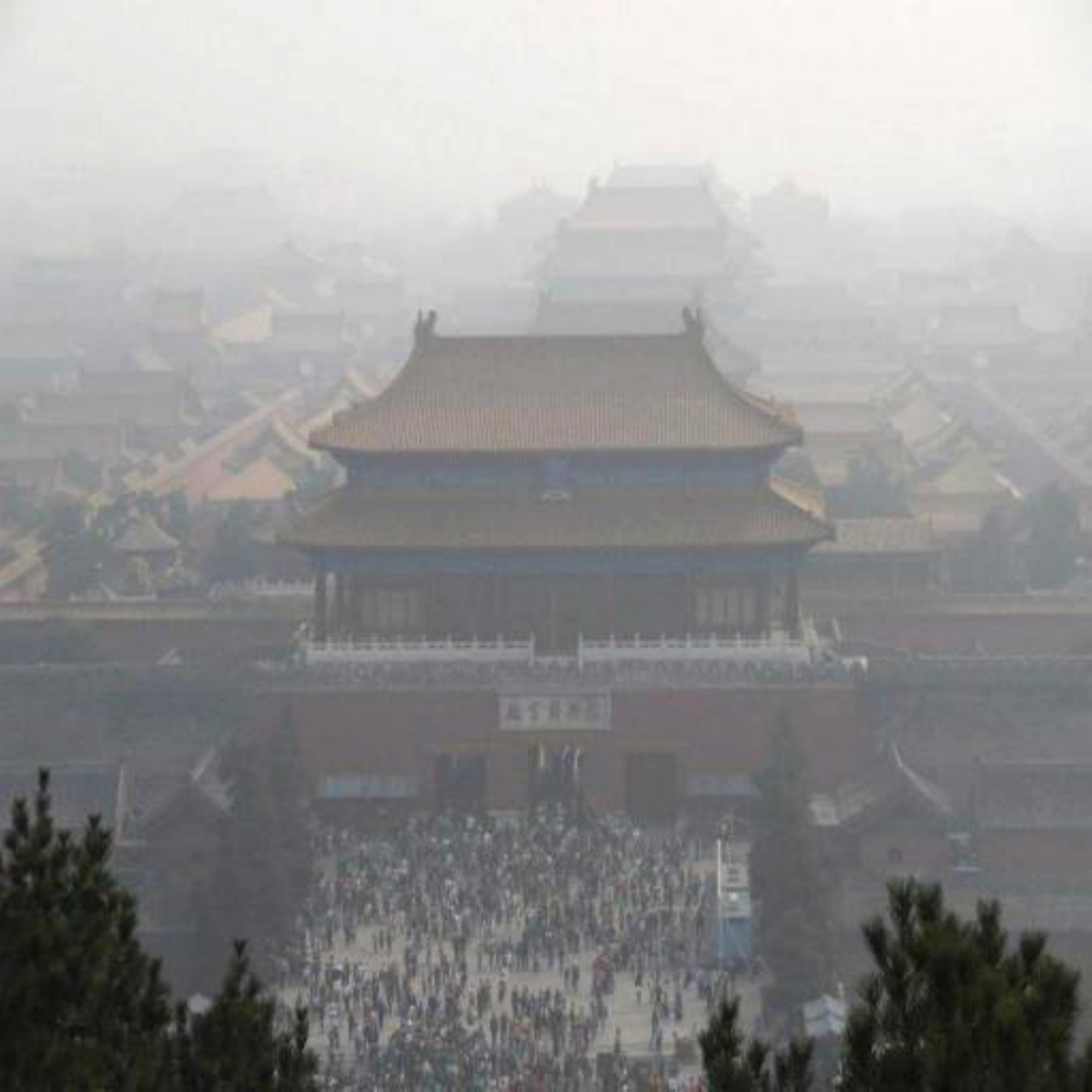}&
\includegraphics[width=0.16\linewidth]{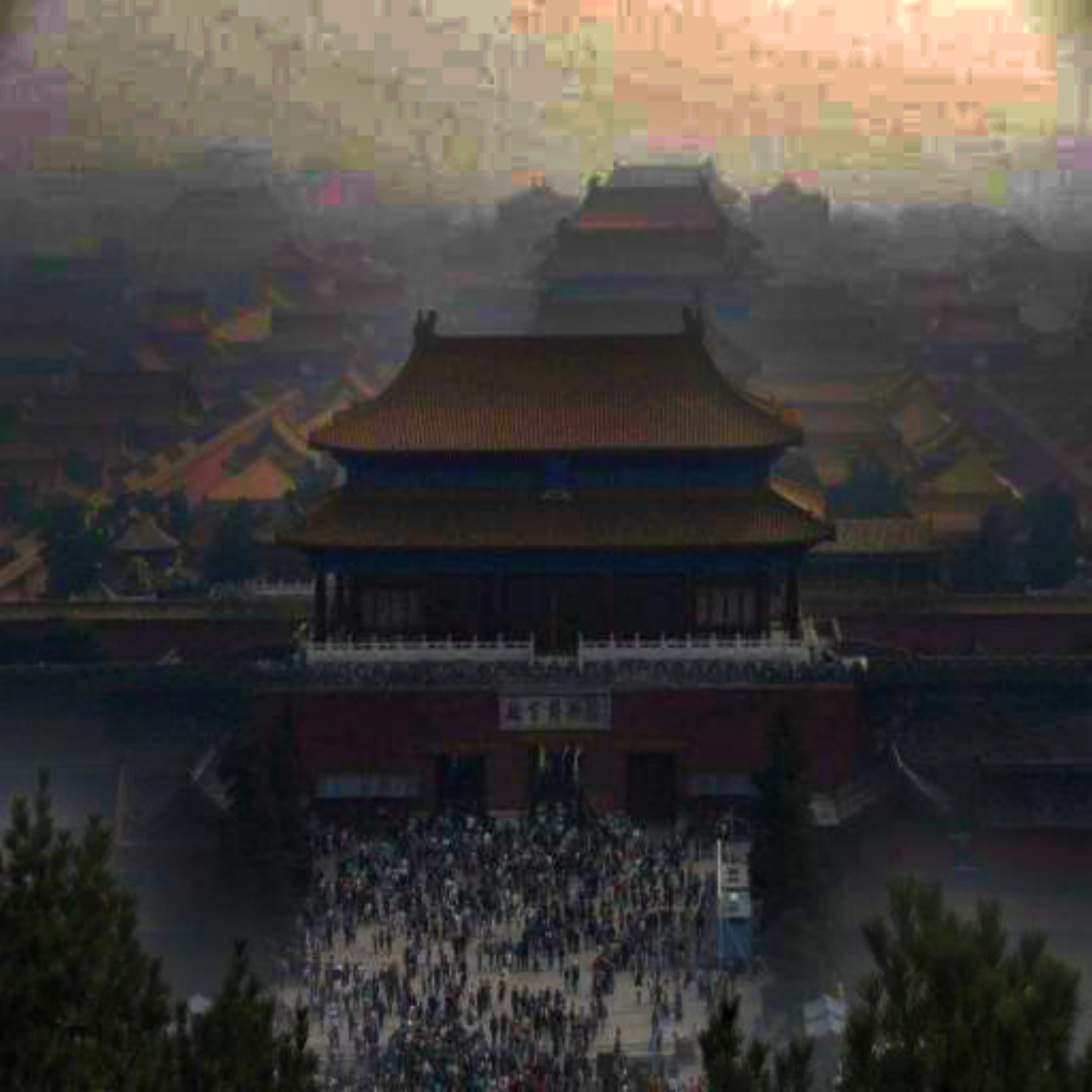}&
\includegraphics[width=0.16\linewidth]{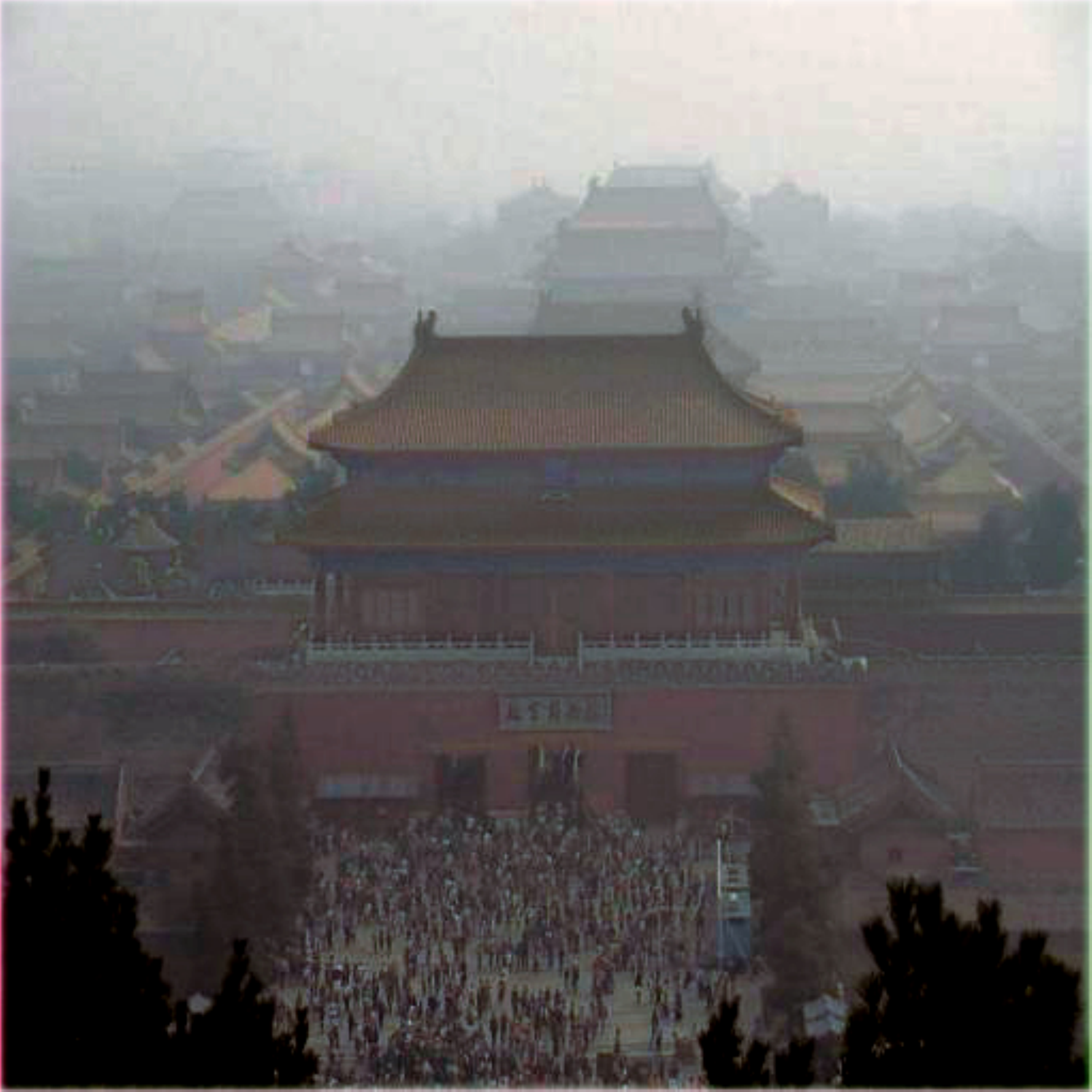}&
\includegraphics[width=0.16\linewidth]{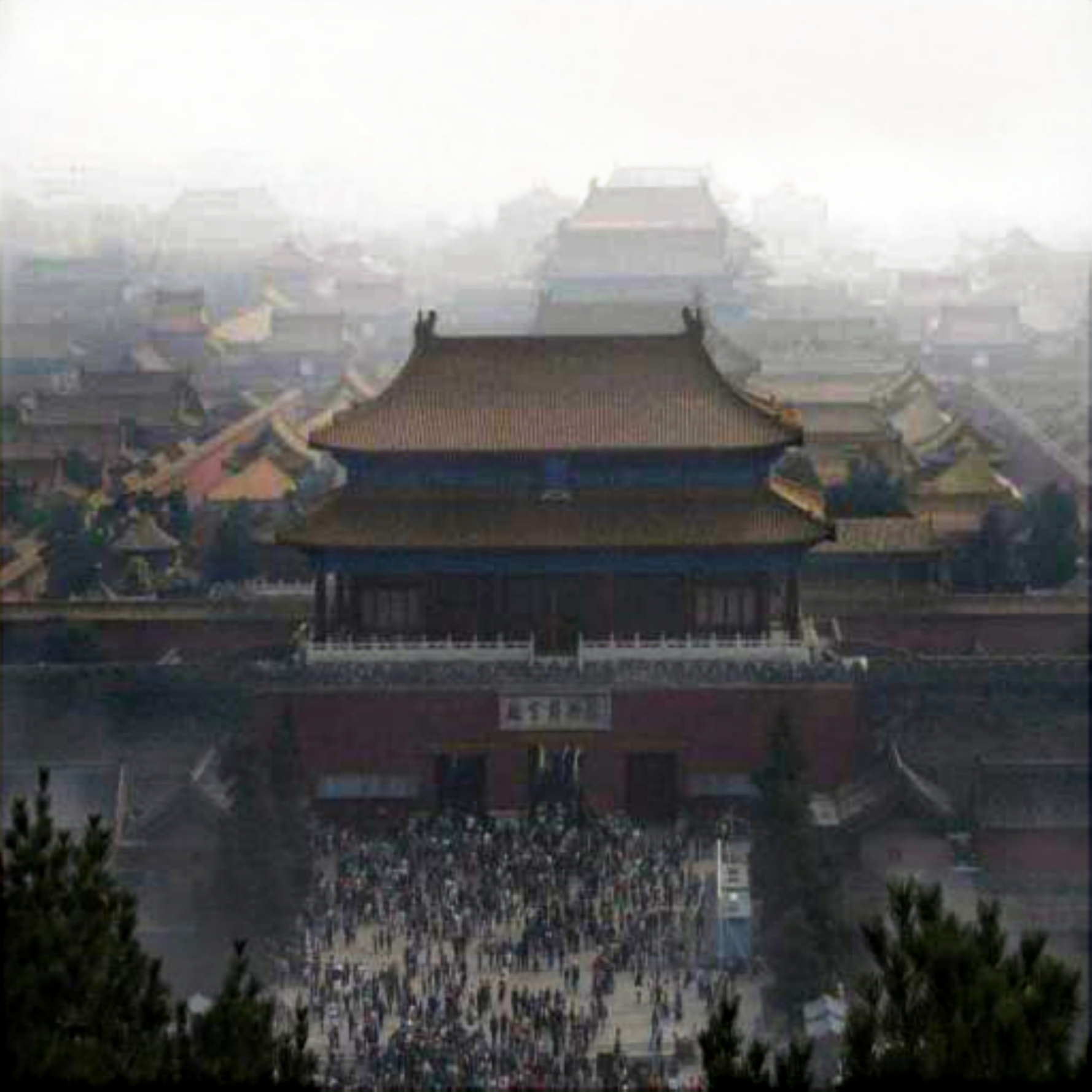}&
\includegraphics[width=0.16\linewidth]{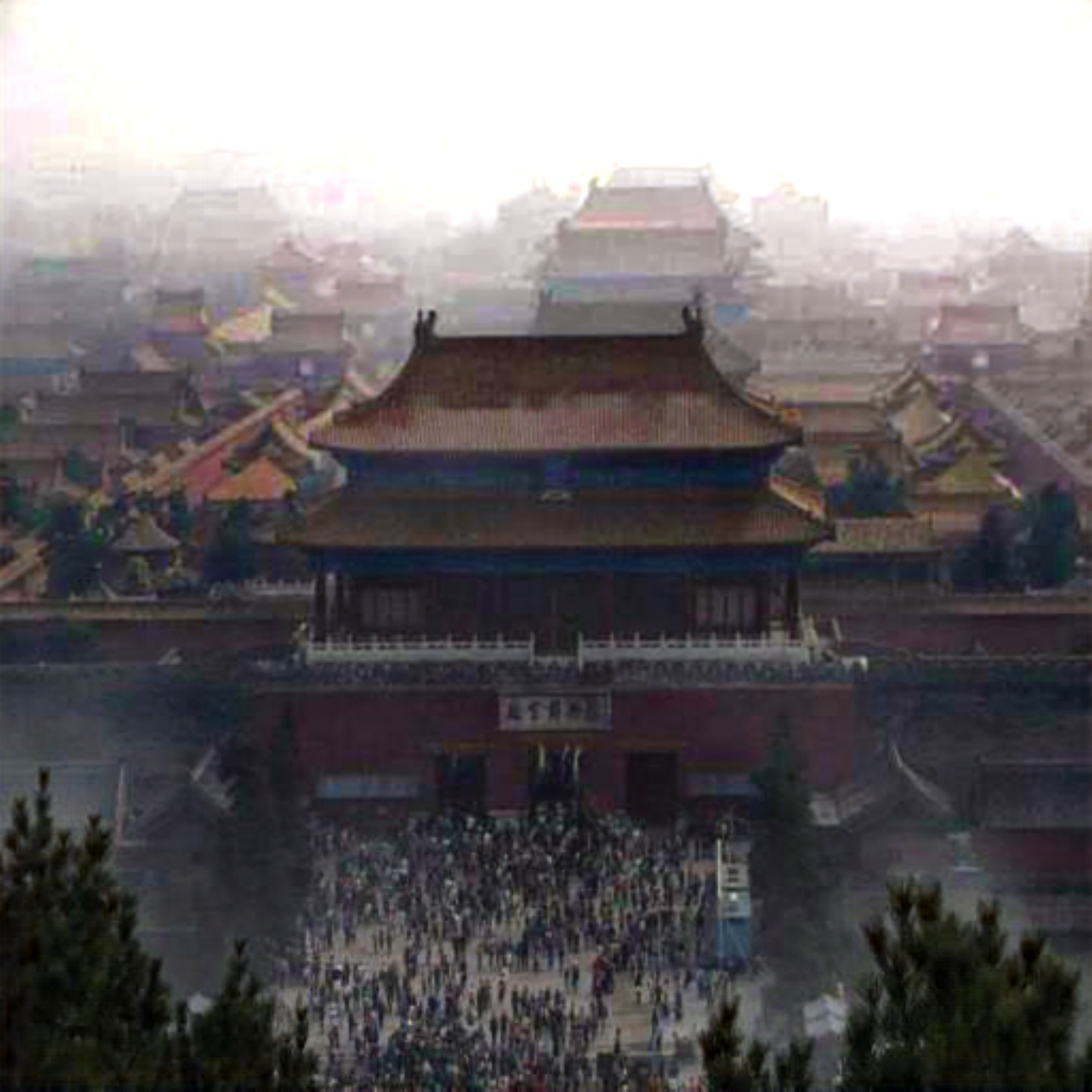}&
\includegraphics[width=0.16\linewidth]{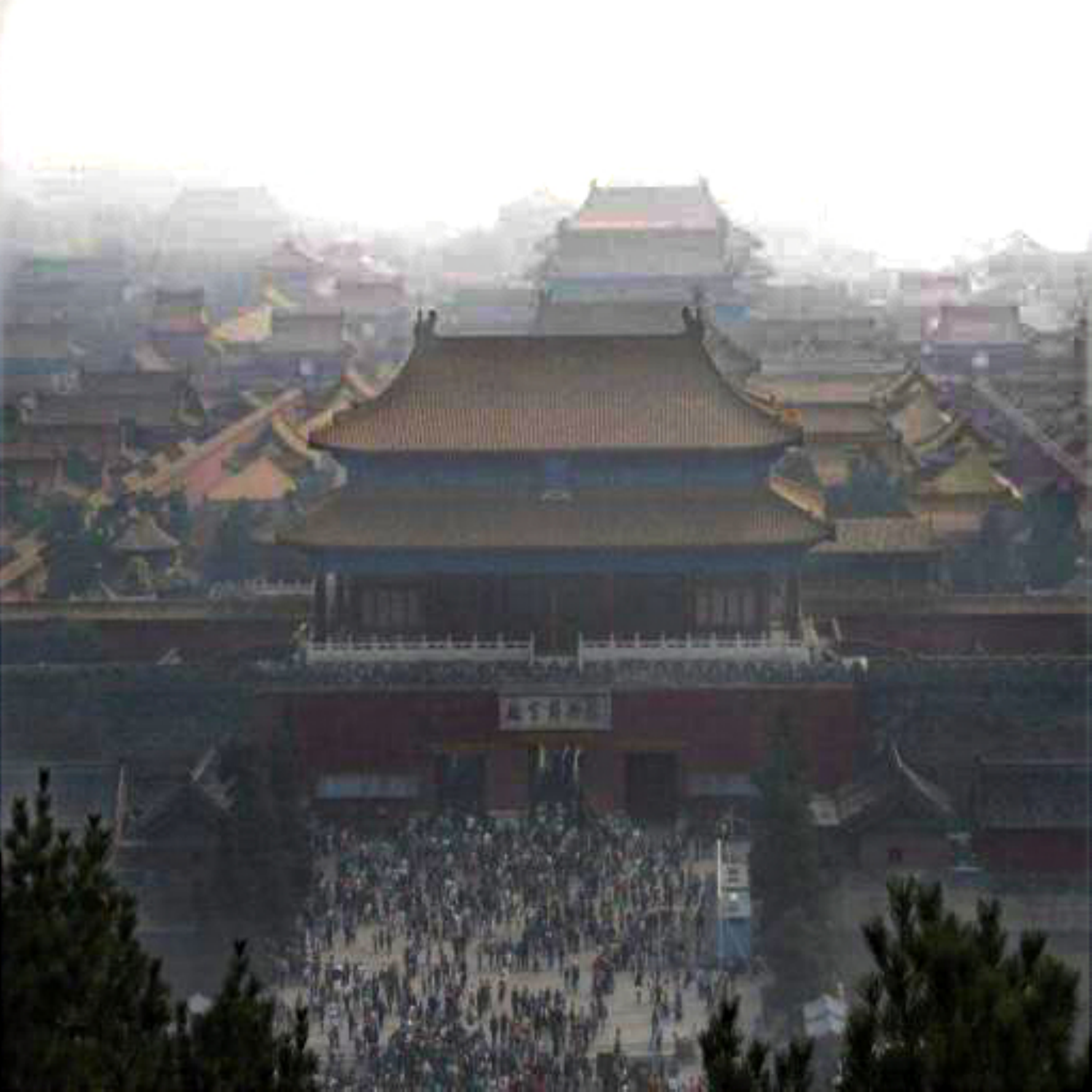}\\

\includegraphics[width=0.16\linewidth]{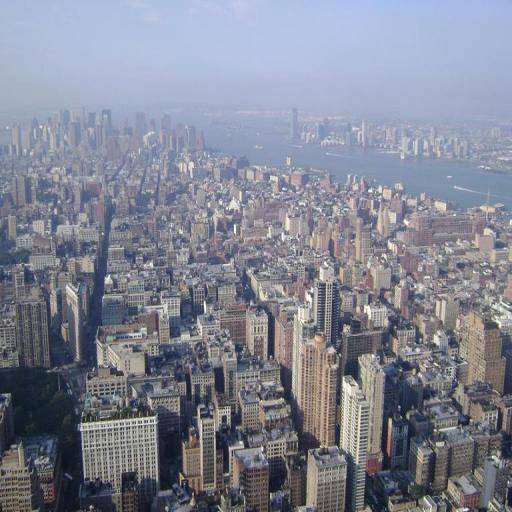}&
\includegraphics[width=0.16\linewidth]{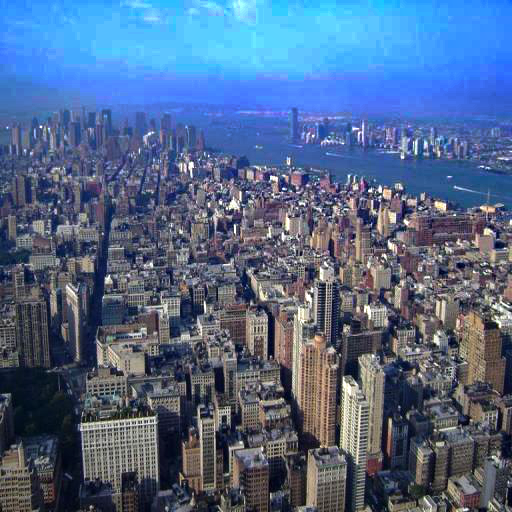}&
\includegraphics[width=0.16\linewidth]{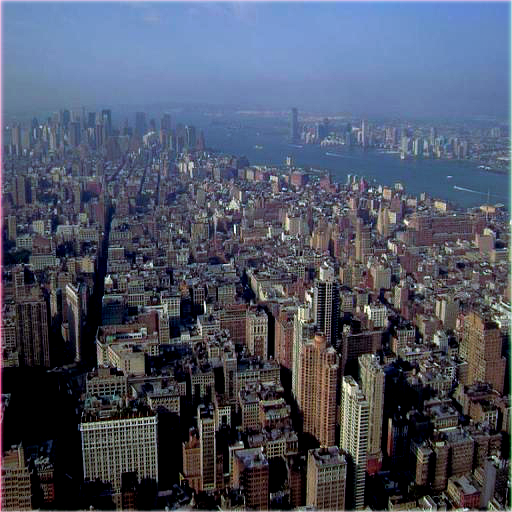}&
\includegraphics[width=0.16\linewidth]{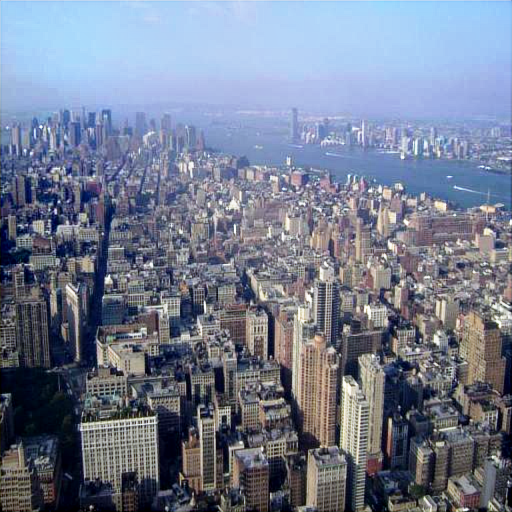}&
\includegraphics[width=0.16\linewidth]{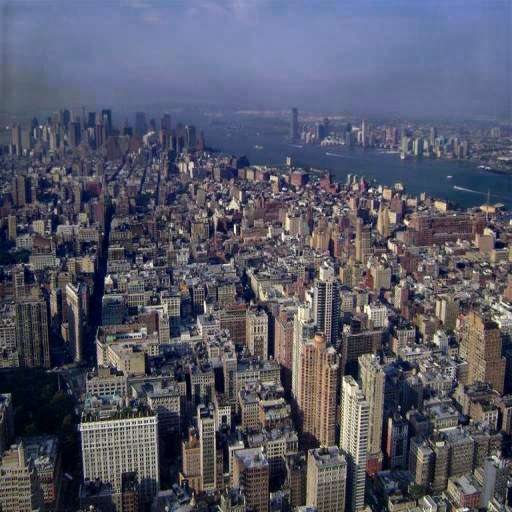}&
\includegraphics[width=0.16\linewidth]{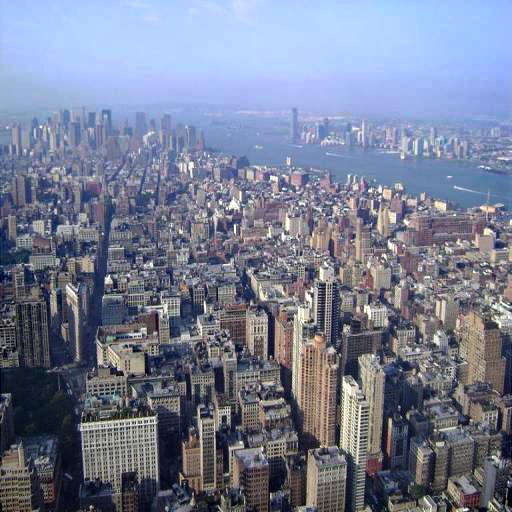}\\

 Input & DCP~\cite{he2010single} &AOD-Net~\cite{li2017aod} &DCPDN~\cite{zhang2018densely}  &GCANet~\cite{chen2018gated}  & PDLD(ours) \\
\end{tabular}
}
\caption{Dehazing comparisons for the real images.}
  \label{fig:real}
\end{figure*}

\begin{table*}[ht]
  \caption{Quantitative comparisons for depth estimations on KITTI dataset.}
  \vspace{-2mm}
  \label{tab:compD}
  \scriptsize
  \begin{tabular}{cccccccc}
    \toprule
                       & $\delta_1\uparrow$ &  $\delta_2\uparrow$& $\delta_3\uparrow$  &$rel\downarrow$   &$Sq.rel\downarrow $& $rms\downarrow$  & $log_{10}\downarrow$\\
    \midrule
   DenseDepth+hazy-free & 0.8976  &0.9695 &0.9870 &0.0965 &0.5724 &4.1834 &0.1659 \\
   DenseDepth+hazy      & 0.8227  &0.9335 &0.9719 &0.1275 &1.0291 &6.2307 &0.2223\\
   DenseDepth-FT        & 0.8768  &0.9662 &0.9863 &0.1122 &0.6079 &4.3439 &0.1803\\
   GCANet+DenseDepth    & 0.8638  &0.9518 &0.9801 &0.1097 &0.8036 &5.3752 &0.1936\\
   DCPDN +DenseDepth    & 0.8462  &0.9417 &0.9750 &0.1176 &0.9335 &5.9481 &0.2097\\
   PDLD-d1              & 0.9022  &0.9709 &0.9874 &0.0944 &0.5228 &3.9861 &0.1648\\
   PDLD-d2              & 0.9095  &0.9736 &0.9885 &0.0974 &0.4997 &3.5927 &0.1593\\
    \bottomrule
  \end{tabular}
\end{table*}
\begin{table}[ht]
  \caption{Quantitative performance comparisons between our progressive learning approach and baseline model(only having one stage).}
  \vspace{-2mm}
  \label{tab:progressive}
  \scriptsize
  \begin{tabular}{ccccc}
    \toprule
               & depth estimation &  transmission & dehazed images \\
               &  sq.rel/rmse/$\delta_1$        & PSNR/SSIM & PSNR/SSIM         \\
    \midrule
   Stage 1     &  0.52283/3.98611/0.90219    &  N/A   &   N/A       \\
   Stage 2     &  0.49968/3.59270/0.90947    &    29.15/0.9776  &  25.12/0.9353     \\
  baseline     &  0.58548/4.27984/0.88406 &      26.09/0.9733   &  24.75/0.9082   \\
    \bottomrule
  \end{tabular}
  \vspace{-3mm}
\end{table}

\subsection{Ablation Study}

\noindent \textbf{Comparison with sequentially dehazing and depth estimations.}
Our approach achieves better dehazing and depth estimation performance for hazy images. In this part, our approach is compared with the following configurations on the KITTI test set:

1) Removing haze first and then estimate the scene depth for the dehazed images (`GCANet+DenseDepth' and `DCPDN +DenseDepth').

2) Fine-tuning the state-of-the-art depth estimation models for the hazy images (`DenseDepth-FT').

3)Depth estimation by the state-of-the-art models for the corresponding haze-free images (`DenseDepth+haze-free').

4)Depth estimation by our progressive depth learning approach (`PDLD-d1',`PDLD-d2'), where `PDLD-d1' and `PDLD-d2' are estimated results by our depth estimation module in the first and second stages respectively.

As shown in Fig.\ref{fig:comDepth}, removing haze first and then estimating depth or fine-tuning the state-of-the-art depth estimation models only marginal improves the performance of the depth estimation.

The error metrics in Table \ref{tab:compD} are commonly applied for the depth estimation evaluations, which are threshold accuracy (${\delta}_i$), average relative error (rel), square relative error ($Sq.rel$), root mean squared error (rms), average $log_{10}$ error ($log_{10}$). Please refer to \cite{Alhashim2018} for the detail calculation formula. The $\uparrow$ indicates the larger the values, the better performance, while the $\downarrow$ is to the opposite.

In Fig. \ref{fig:comDepth}, average pixel-wise abs errors are investigated for the pixels whose depths are within certain distances range $[d-2,d]$ in the horizontal axis in KITTI dataset. DenseDepth+hazy-free denotes the DenseDepth model performs on the hazy-free images. The DenseDepth-FT denotes the model fine-tuned for hazy images, PDLD-d1 and PDLD-d2 denote the depth estimation modules in the first and second stages, respectively.

Observed from Table \ref{tab:compD} and Fig.\ref{fig:comDepth}, it is obvious that haze severely disturbs the depth estimation of the existing depth estimation models trained on the hazy-free images. Fine-tuning the existing models of haze-free image depth estimation or combinations of the dehazing and depth estimation sequentially improves the situations but is far from enough. Our progressive depth learning for the dehazing has largely improved the situations. Even the estimated depth in the first stage has outperformed all the above approaches.
It is surprising that our progressively depth learning approach even outperforms depth estimations by the state-of-the-art models for the corresponding hazy-free images. It is obvious that our model captures the inner relationship of depth and transmission and hazes provide some cues for depth estimations. Simply combination of dehazing and depth estimation or simple adaptation like fine-tuning provide marginal help for the situations.

\subsection{Progressive Depth Learning }
In this part, the depth estimations and dehazing results by each stage are compared on KITTI dataset.
The baseline denotes our backbone model with only one stage. It is clear in table \ref{tab:progressive} that our progressive learning approach largely surpasses the dehazing and depth estimation performances of baseline model. It is interesting that with our progressive learning strategies, the depth estimation module in the first stage is greatly improved. The depth estimation module in the first stage benefits from the improved transmission estimation module and more supervision is back-propagated from the progressive learning process. Depth estimation by the second stage produces more accurate results than the first stage which proves the training progressively refine the estimations. Besides, guidance by more accurate depth estimations largely improve the transmission estimation and dehazing results which clearly proves the power of exploiting the relationship between dehazing and depth estimation. The dehazing and depth estimation forms a positive feedback and the progressive depth learning and dehazing approach improves each term iteratively.

\subsection{Evaluations on the Real Dataset}
To demonstrate the effectiveness on the real applications, the proposed approach is evaluated on real word hazy images provided by the existed work~\cite{zhang2018densely}. As revealed in Fig.\ref{fig:real}, method of DCP~\cite{he2010single} has the problem in the regions where the light is very bright,such as sky. AOD-Net~\cite{li2017aod} and GCANet~\cite{chen2018gated} seems to make some regions dark and leave some hazes. The contrasts is too large in the second row for \cite{zhang2018densely}. The hazy removal of \cite{zhang2018densely} is not uniform which looks unnatural. Our approach produces visually pleasing dehazed images.

\section{Conclusions}
In this paper, a progressive depth learning strategy for single image dehazing is proposed to initialize depth estimations with the help of the transfer learning from the hazy-free image depth estimation domains and iteratively refine the depth and transmission estimation. Our approach exploits the inner relationships between image depth and transmission which not only alleviate the under-constraint problem of the transmission estimations and largely improve the dehazing performances, but also achieve significant more accurate depth estimation results for the hazy images. It is surprisingly sometimes the performance of our model obtains even better depth estimation than the state-of-the-art model performed on the hazy-free images as the hazes provide some cues for the depth estimations. Our approach restores visual pleasing dehazing results for both synthetic and real hazy images. We hope to arouse the interests of the communities to explore the relationships between image depth and dehazing to improve the both tasks.


\bibliographystyle{ACM-Reference-Format}
\bibliography{haze}

\end{document}